\documentclass[10pt,twocolumn,letterpaper]{article}

\usepackage[dvipsnames]{xcolor} 
\usepackage{cvpr}              %

\usepackage{physics}
\usepackage{amsmath}
\usepackage[cjk]{kotex} %
\usepackage{multirow}
\usepackage{booktabs}
\usepackage{tabularx}
\usepackage{colortbl}
\usepackage{graphicx} %
\usepackage{bbding}
\usepackage{pifont}
\usepackage{wasysym}
\usepackage{amssymb}
\usepackage{pdfrender}
\usepackage{wrapfig} %
\usepackage{subcaption}
\usepackage{caption}
\usepackage{scalerel}
\usepackage{algorithm}
\usepackage{algpseudocodex}
\usepackage{bbm} %
\usepackage{xcolor}
\usepackage{kotex} 
\usepackage{soul}
\usepackage{array}

\newcommand{\mycc}{\cellcolor{gray!20}}

\newcolumntype{R}{>{\columncolor{gray!20}}c}

\newcommand\clearrow{\global\let\rowmac\relax}
\newcommand{\specialcell}[2][c]{%
  \begin{tabular}[#1]{@{}c@{}}#2\end{tabular}}

\newcommand*{\boldcheckmark}{%
  \textpdfrender{
    TextRenderingMode=FillStroke,
    LineWidth=.5pt, %
  }{\checkmark}%
}
\clearrow

\newcommand\ourssl{view-batch}

\usepackage{mathtools}
\usepackage{amsthm}
\usepackage[accsupp]{axessibility}

\definecolor{cvprblue}{rgb}{0.21,0.49,0.74}
\usepackage[pagebackref,breaklinks,colorlinks,citecolor=cvprblue,linkcolor=BrickRed,urlcolor=PineGreen,hypertexnames=false]{hyperref}
\usepackage{url}
\usepackage[capitalize,noabbrev]{cleveref}

\def\paperID{****} %
\def\confName{CVPR}
\def\confYear{2025}

\title{
Do Your Best and Get Enough Rest for Continual Learning
}

\author{Hankyul Kang$^1$
\and
Gregor Seifer$^1$
\and
Donghyun Lee$^2$
\and
Jongbin Ryu$^{1}$ \thanks{Corresponding author.} \and \\ 
$^1$ Ajou University \, $^2$ KAIST
}

\begin{document}
\maketitle

\begin{abstract}
According to the forgetting curve theory, we can enhance memory retention by learning extensive data and taking adequate rest.
This means that in order to effectively retain new knowledge, it is essential to learn it thoroughly and ensure sufficient rest so that our brain can memorize without forgetting. 
The main takeaway from this theory is that learning extensive data at once necessitates sufficient rest before learning the same data again.
This aspect of human long-term memory retention can be effectively utilized to address the continual learning of neural networks.
Retaining new knowledge for a long period of time without catastrophic forgetting is the critical problem of continual learning.
Therefore, based on Ebbinghaus' theory, we introduce the view-batch model that adjusts the learning schedules to optimize the recall interval between retraining the same samples.
The proposed view-batch model allows the network to get enough rest to learn extensive knowledge from the same samples with a recall interval of sufficient length. 
To this end, we specifically present two approaches: 
1) a replay method that guarantees the optimal recall interval, and 2) a self-supervised learning that acquires extensive knowledge from a single training sample at a time.
We empirically show that these approaches of our method are aligned with the forgetting curve theory, which can enhance long-term memory.
In our experiments, we also demonstrate that our method significantly improves many state-of-the-art continual learning methods in various protocols and scenarios.
\urlstyle{rm}
We open-source this project at \url{https://github.com/hankyul2/ViewBatchModel}.
\end{abstract}

\section{Introduction}
The forgetting curve theory~\citep{ebbinghaus2013memory} suggests that human memory tends to fade as time goes on; however, memory retention can be improved by repeated learning in an optimal recall interval.
Motivated by this theory, many subsequent studies~\citep{gordon1925class,cepeda2006distributed,kang2016spaced} have shown that spaced repetition with optimal recall interval enhances long-term memory retention in the human brain. 
This phenomenon is known as the spacing effect~\citep{cepeda2008spacing}, where the optimal recall interval for repeated learning is crucial to maintain an acceptable level of forgetting knowledge. %
Several studies~\citep{shaughnessy1977long,cepeda2008spacing,hawley2008comparison} have made significant progress on the spacing effect, shedding light on its effectiveness.
There have also been studies~\citep{mace1932psychology,melton1970situation,karpicke2007expanding} highlighting the importance of the recall interval of repeated learning.

\begin{figure}[t]
    \centering

    \hfill
    \hfill
    \begin{subfigure}[b]{0.47\linewidth}
        \centering        
        \includegraphics[width=\linewidth]{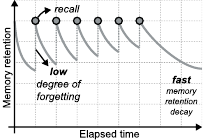}
        \caption{Short-term recall interval}
        \label{fig:f_forget_curve_short}
    \end{subfigure}
    \hfill
    \begin{subfigure}[b]{0.47\linewidth}
        \centering        
        \includegraphics[width=\linewidth]{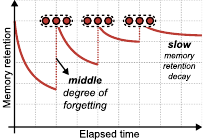}
        \caption{Optimal recall interval}
        \label{fig:f_forget_curve_optimal}
    \end{subfigure}
    \hfill\hfill\hfill

    \hfill\hfill
    \begin{subfigure}[b]{0.47\linewidth}
        \centering        
        \includegraphics[width=\linewidth]{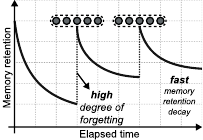}
        \caption{Long-term recall interval}
        \label{fig:f_forget_curve_long}
    \end{subfigure}
    \hfill
    \begin{subfigure}[b]{0.47\linewidth}
        \centering        
        \includegraphics[width=\linewidth]{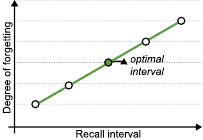}
        \caption{Degree of forgetting}
        \label{fig:f_degree_of_forgetting}
    \end{subfigure}
    \hfill
    \hfill
    \hfill
    
    \caption{ %
    \textbf{Conceptual graph of the forgetting curve.} We show (a) short-term recall interval, (b) optimal recall interval, (c) long-term recall interval, and (d) degree of forgetting. 
    (a-b) Expanding the recall interval improves long-term memory retention of neural networks by repeatedly recalling memory with moderate difficulty, whereas (c) an excessive recall interval decreases it. 
    The depicted forgetting curve regarding recall interval is based on the spacing effect formula~\citep{cepeda2006distributed,cepeda2008spacing} provided in the supplementary material.
    }
    \label{fig:f_forget_curve}
\end{figure}

\begin{figure}[t]
    \centering

    \hfill
    \hfill
    \includegraphics[width=.4\linewidth]{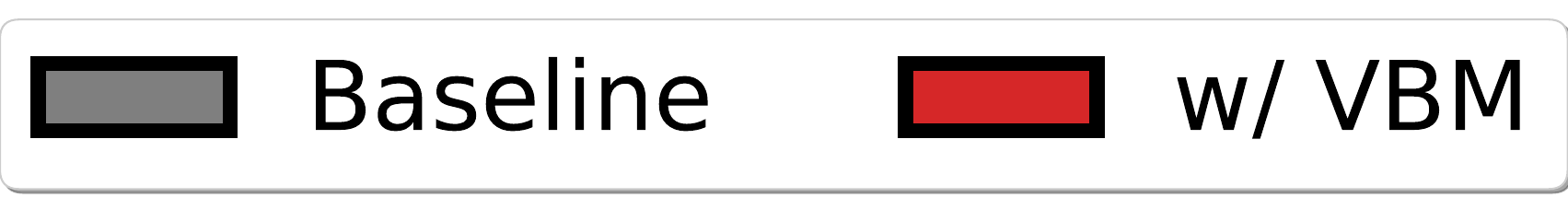}
    \hfill
    \hfill
    \hfill

    \hfill
    \hfill
    \begin{subfigure}[t]{0.32\linewidth}
        \centering        
        \includegraphics[width=\textwidth]{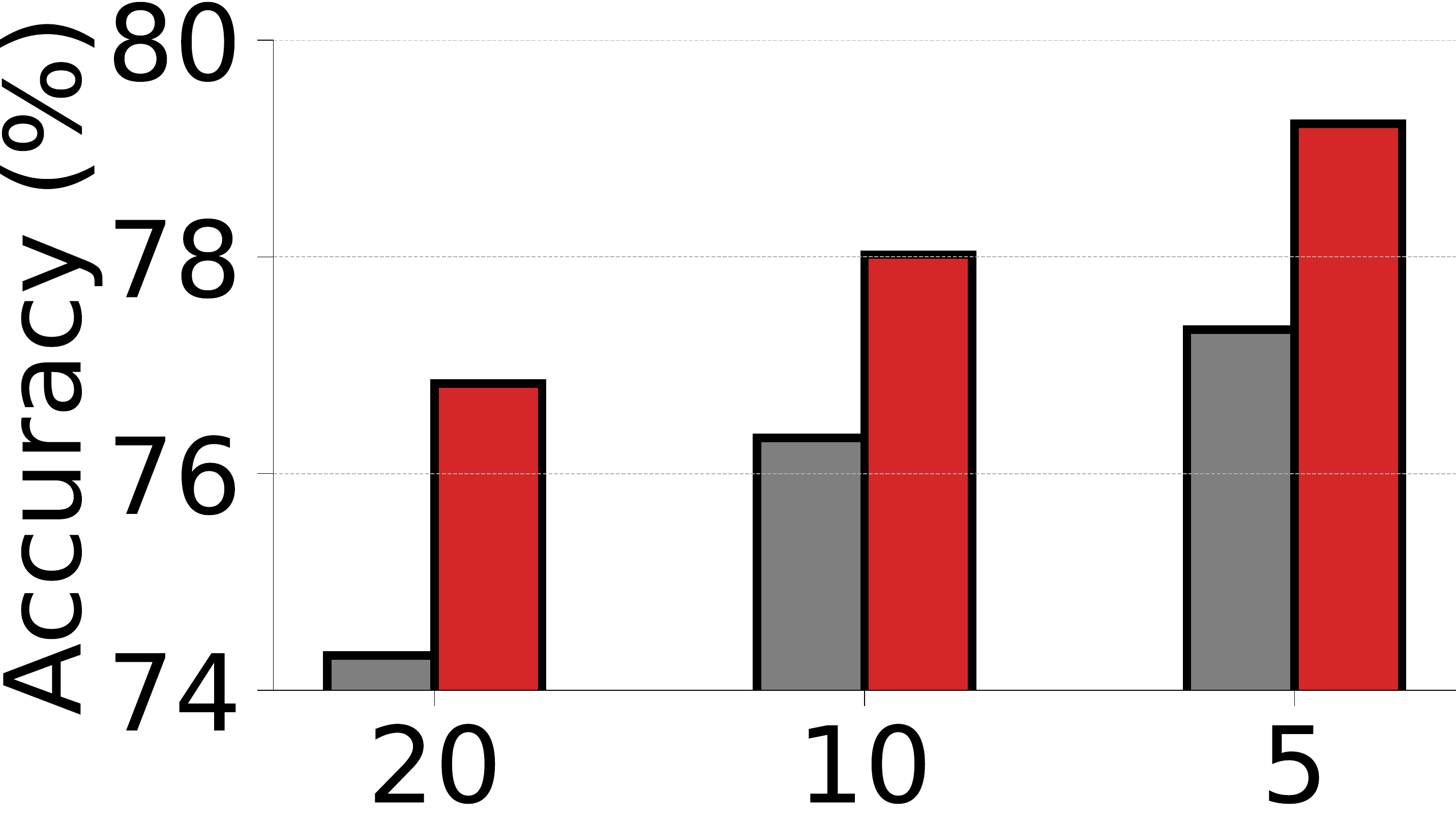}
        \caption{Step size}
        \label{fig:f_po1}
    \end{subfigure}
    \hfill
    \begin{subfigure}[t]{0.32\linewidth}
        \centering        
        \includegraphics[width=\textwidth]{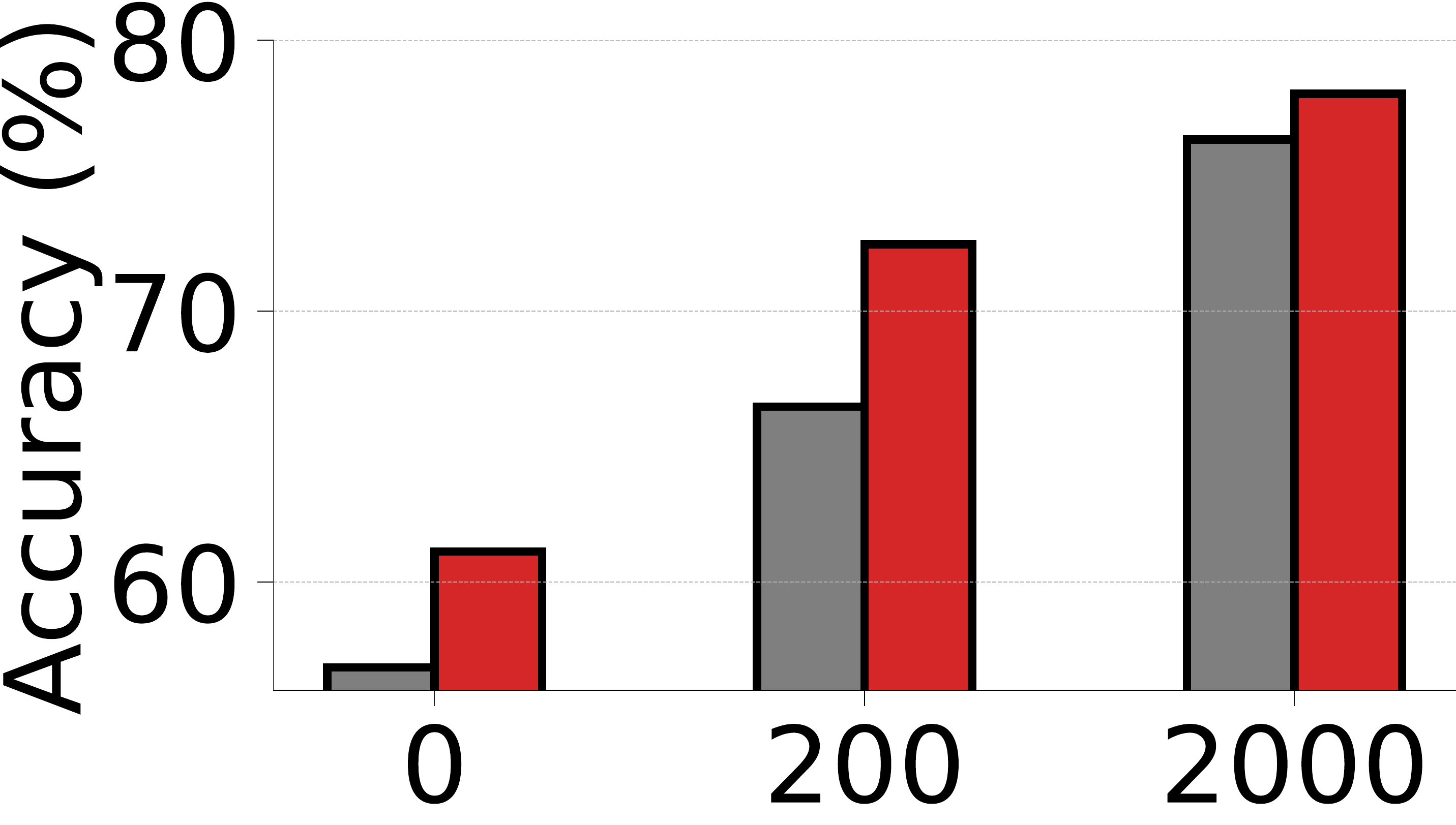}
        \caption{Buffer size}
        \label{fig:f_po2}
    \end{subfigure}
    \hfill
    \begin{subfigure}[t]{0.32\linewidth}
        \centering        
        \includegraphics[width=\textwidth]{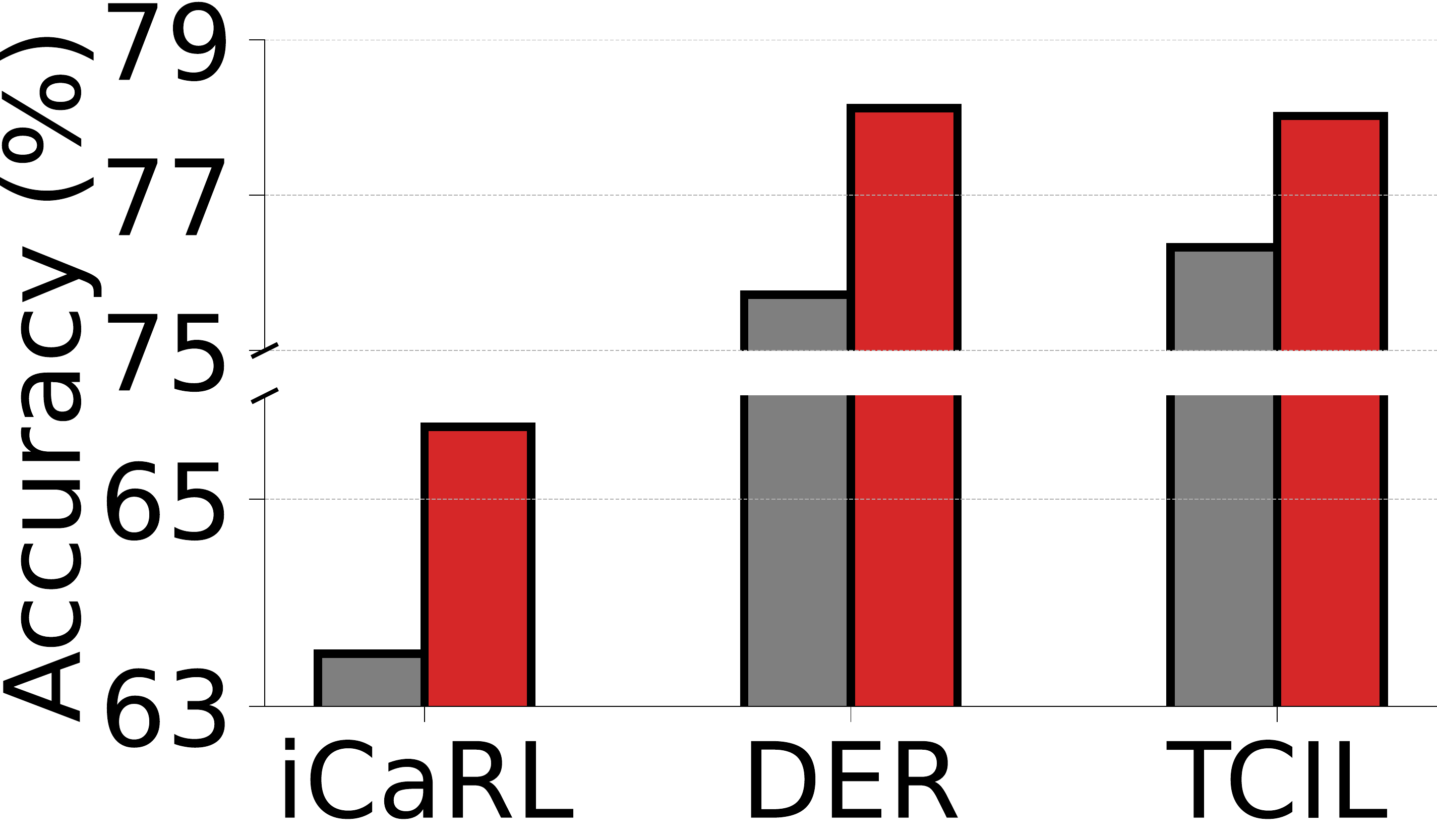}
        \caption{Baseline method}
        \label{fig:f_po3}
    \end{subfigure}
    \hfill\hfill\hfill
    
    \hfill\hfill
    \begin{subfigure}[t]{0.32\linewidth}
        \centering        
        \includegraphics[width=\linewidth]{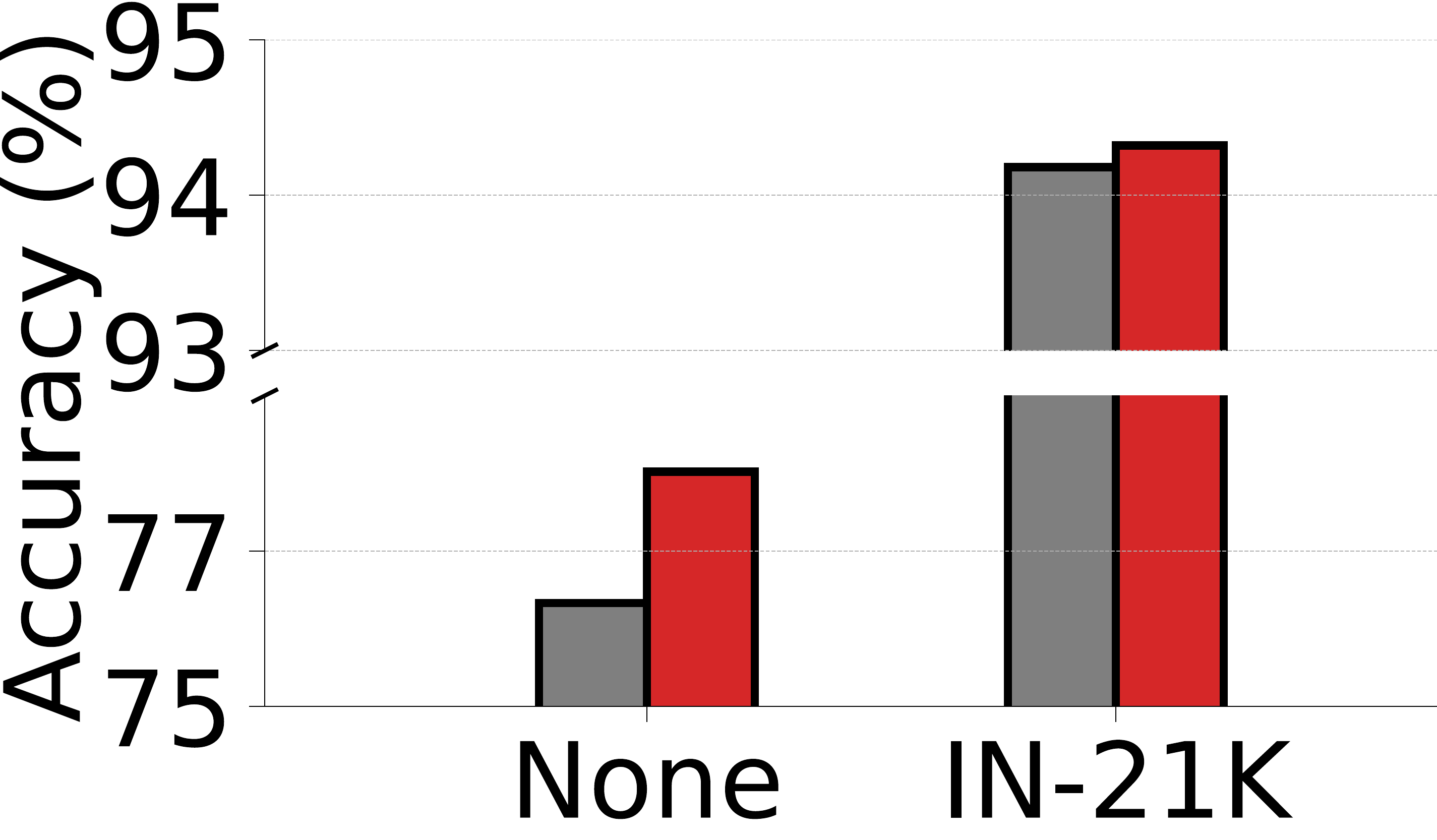}
        \caption{Pre-trained model}
        \label{fig:f_po4}
    \end{subfigure}
    \hfill
    \begin{subfigure}[t]{0.32\linewidth}
        \centering        
        \includegraphics[width=\linewidth]{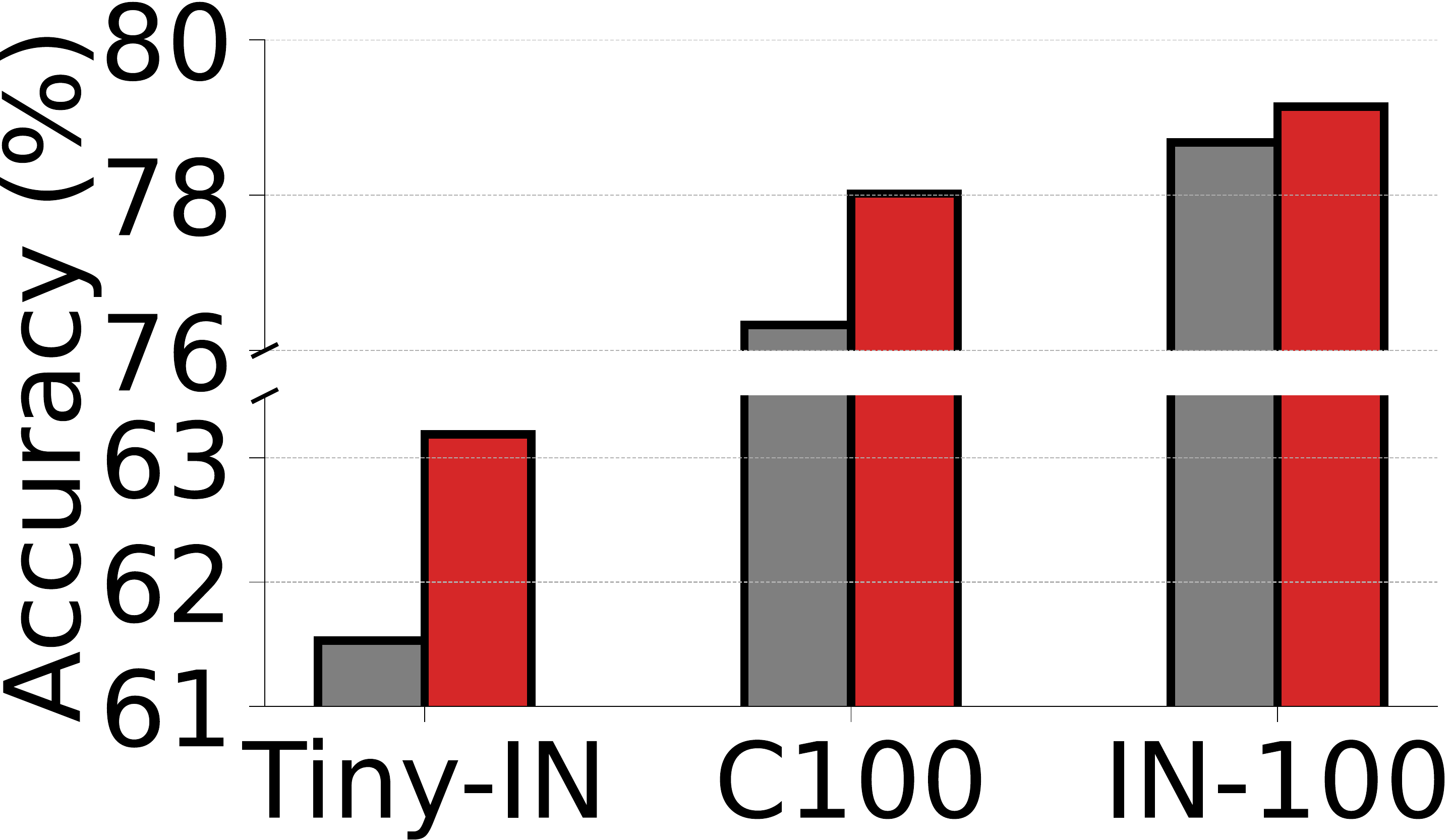}
        \caption{Benchmark}
        \label{fig:f_po5}
    \end{subfigure}
    \hfill
    \begin{subfigure}[t]{0.32\linewidth}
        \centering        
        \includegraphics[width=\linewidth]{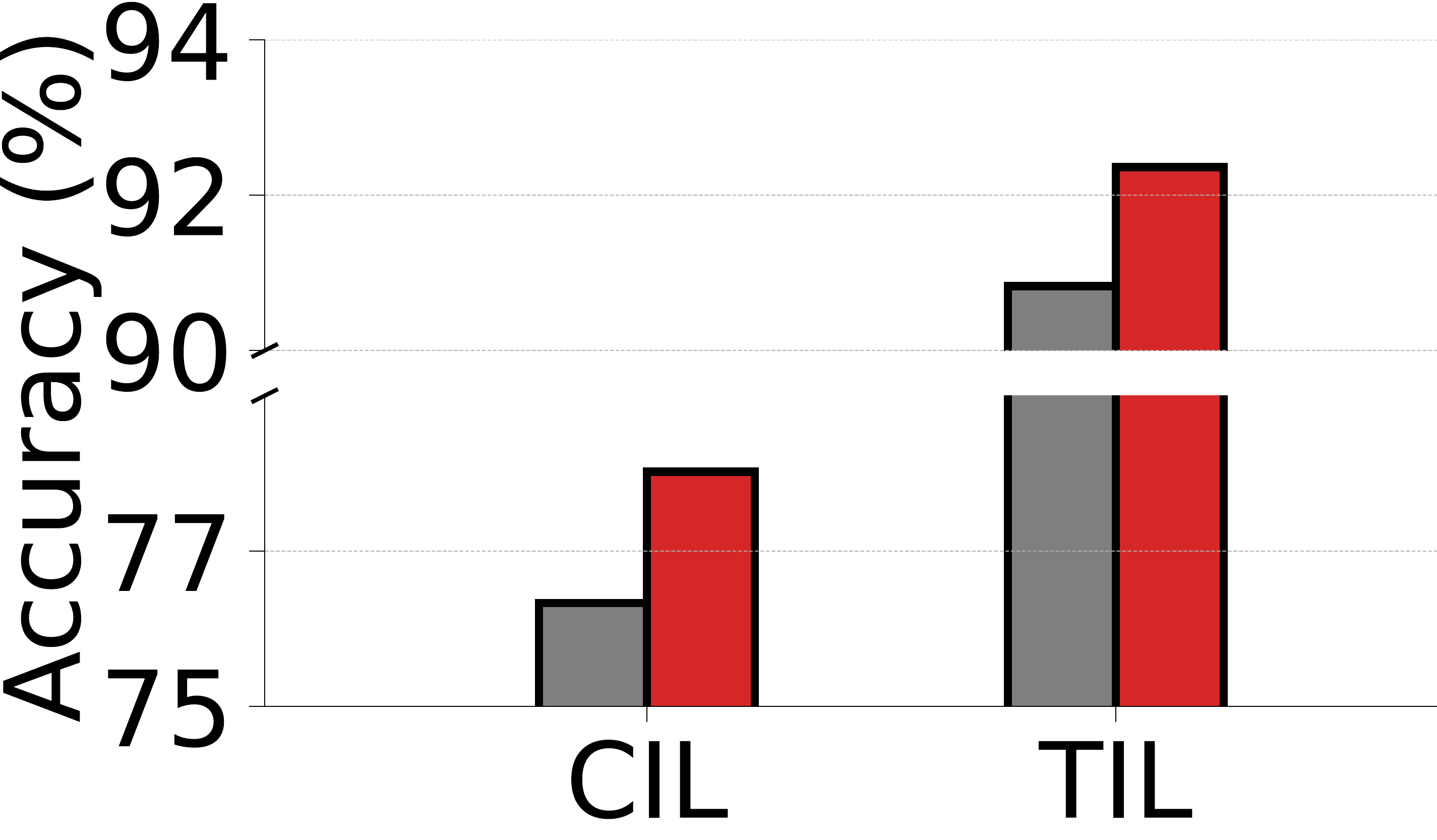}
        \caption{Protocol}
        \label{fig:f_po6}
    \end{subfigure}
    \hfill
    \hfill
    \hfill
    
    \caption{
    \textbf{Overview of experimental results.}
    We provide comprehensive comparisons of various factors for continual learning. We perform extensive experiments on step and buffer sizes (a-b), three different continual learning methods (c), whether to use the pre-trained model (d), three different benchmarks (e), and two evaluation protocols (f). In all cases, ours improves the baseline performance consistently.
    }
    \label{fig:plot_po}
\end{figure}

Although this theory has been widely recognized in human brain research, it is currently very little known in the field of machine learning. %
Only a few works~\citep{amiri-etal-2017-repeat,cao-etal-2024-retentive-forgetful,zhong2024memorybank} show this theory's potential in neural networks by utilizing it as an analysis toolkit or dynamic learning strategy.
So, we investigate the impact of the forgetting curve theory on long-term memory retention in neural networks for continual learning.

In Ebbinghaus' theory, we focus on the decay of memory retention in the forgetting curve as shown in~\cref{fig:f_forget_curve}. 
The forgetting curve suggests that, as shown in~\cref{fig:f_forget_curve_short}, the sample memory shows extreme knowledge decay when repeating its learning with a short time-space. On the other hand, as shown in~\cref{fig:f_forget_curve_long}, if there is an excessive recall interval of repeated learning, too much memory is forgotten, leading to a decreased level of memory retention when relearning the same sample.
However, as shown in~\cref{fig:f_forget_curve_optimal}, when there is an optimal recall interval between repeated learning, the decay of memory retention also becomes gentle. 
This effect of the optimal recall interval implies that it is imperative to implement an appropriate learning schedule to retain memories efficiently in the repeated learning process. 
The notion behind this spacing effect can be usefully applied in continual learning, particularly to prevent catastrophic forgetting.

However, empirically, we found that the current methods for continual learning have a short recall interval~\citep{rebuffi2017icarl,huang2023resolving}, %
so they are limited to profit from the spacing effect, which requires a sufficient recall interval.

To tackle this limitation, we introduce two approaches for training neural networks.
Our first approach aims to prevent short recall intervals through the proposed replay method.
We replay {\ourssl} in the repeated training schedule so that short recall intervals can be delayed sufficiently.
We augment multiple views from a single sample, and thus, the recall intervals of repeated learning can be adjusted.
In our second approach, we make use of self-supervised learning to compensate for the delayed recall interval by training each sample extensively.
We leverage the one-to-many divergence to learn the generalized self-supervised features within a view-batch, which enhances the capability of the networks.
These two approaches ensure that the neural networks have enough time to retrain the same samples, and at the same time, they can acquire substantial knowledge from training samples at a time.
We apply our view-batch model (VBM) to both rehearsal and rehearsal-free continual learning scenarios, where addressing long-term memory retention is critical.
Moreover, we show extensive experimental results of various continual protocols and scenarios. 
As an overview, \cref{fig:plot_po} showcases the comprehensive comparison against different step sizes, memory buffer size, baseline methods, use of the pre-trained model, benchmarks, and protocols.
For all these comparisons, our view-batch model achieves significant performance improvements. 
We have condensed our contributions into the following points: 
\begin{itemize}
\item We propose the view-batch model that optimizes the recall interval between retraining samples in order to enhance memory decay in the continual learning task. The proposed view-batch model includes replay and self-supervised learning components to accomplish extensive learning in optimized intervals.
\item We showcase the effectiveness of our approach on various state-of-the-art continual learning methods, where ours consistently improves the performance. %
The proposed method is a drop-in replacement approach, so it does not require any extra computational costs for the performance improvements. 
\item We reveal that enhancing memory decay can significantly improve the performance of the continual learning. This confirms the practical utility of theoretical knowledge for neural networks.
\end{itemize}

\section{Related Work}

\subsection{Forgetting Curve Theory in Neural Networks}
Several studies~\citep{kali2002replay,hunziker2019teaching} have employed the forgetting curve theory~\citep{ebbinghaus2013memory} to gain insights into the inherent memory retention of neural networks. 
Recently, \citet{tirumala2022memorization} investigated the forgetting curve theory in neural networks regarding different model scales.
\citet{cao-etal-2024-retentive-forgetful} have also utilized the forgetting curve theory to examine the variations in memory conformation of the neural networks as influenced by pre-training methods. 
Additionally, \citet{amiri-etal-2017-repeat} proposed a dynamic learning scheduler to reduce total learning cost, leveraging optimal learning intervals corresponding to the forgetting curve.
\citet{zhong2024memorybank} optimized memory retrieval timing based on the principles of the spacing effect theory, enabling the retention of the previous dialogue context existing over an extended duration.

\subsection{Data Augmentation}
Data augmentation~\citep{yun2019cutmix,cubuk2020randaugment} has been widely used in various existing networks.
It helps networks learn rich visual information using a limited number of training samples.
Data augmentation is also effective in self-supervised learning~\citep{caron2021emerging,chen2020simple}.
Unlimited labels can be generated when different augmentation methods are applied to a single sample.
Therefore, the network is able to learn with self-supervised labels of augmented samples.
We use this augmentation method to construct multiple views in a view-batch and also learn self-supervision along with the supervised information of the continual learning process.
There have been several studies~\citep{hoffer2019augment,berman2019multigrain} on repeated augmentation, which allows learning multiple views from a single sample at a time. However, these studies only focused on improving the performance of the augmentation method and did not examine the effect of recall intervals. In addition, they do not apply self-supervised learning and only use label supervision to train networks.

\subsection{Continual Learning}
The main problem of continual learning is preserving old knowledge while acquiring new one.
This problem has been recognized as catastrophic forgetting, and one common strategy used to address this problem is to make use of a memory buffer~\citep{rebuffi2017icarl}.
The approaches to this strategy can be divided into two categories: one involves the direct storage of a subset of training samples from a previously learned task~\citep{rebuffi2017icarl,riemer2018learning}, while the other generates training samples from the learned task~\citep{shin2017continual}.
This approach of utilizing memory buffers proves to be highly effective in addressing the issue of catastrophic forgetting; however, it does come with the trade-off of requiring additional memory space. 
Consequently, non-rehearsal continual learning methods~\citep{riemer2018learning,dhar2019learning} have been widely studied to avoid additional memory usage.
In addition, studies~\citep{wang2022learning,wang2022dualprompt,zhang2023slca} using pre-trained models in non-rehearsal continual learning have been actively conducted recently. 
Meanwhile, there have been several works~\citep{li2017learning,douillard2020podnet} that tackle continual learning from the perspective of the regularization method. The architecture-based method~\citep{serra2018overcoming,yan2021dynamically} has been another prevalent research direction for addressing continual learning.

\begin{figure}[t]
    \centering

    \hfill
    \hfill
    \begin{subfigure}[b]{0.47\linewidth}
        \centering        
        \includegraphics[width=\linewidth]{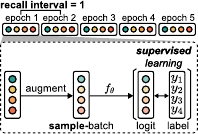}
        \caption{Baseline}
        \label{fig:f_baseline}
    \end{subfigure}
    \hfill
    \begin{subfigure}[b]{0.47\linewidth}
        \centering        
        \includegraphics[width=\linewidth]{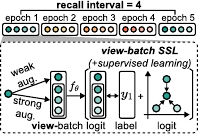}
        \caption{View-batch}
        \label{fig:f_ours}
    \end{subfigure}
    \hfill
    \hfill
    \hfill
    
    \caption{
    \textbf{Schematic illustration of the proposed view-batch model.}
    In subfigure (b), we show our view-batch model employing the replay (V=4) and self-supervised learning approach.
    In contrast to (a) the baseline method, we learn multiple views of the same sample (marked as different shades) using the proposed {\ourssl} self-supervised loss to learn it extensively and ensure enough time-space between recall intervals.
    For simplicity, we assume in (a) that the entire training data and batch size are the same as four, thus a single training epoch constitutes one batch.
    }
    \label{fig:f_learning_strategy}
\end{figure}

\section{Method}

We introduce a view-batch model designed to enhance the long-term memory retention of neural networks for continual learning (CL) tasks. 
In existing CL methods~\citep{rebuffi2017icarl,yan2021dynamically,huang2023resolving}, the neural networks retrain the same samples with short time intervals, which prohibits long-term memory retention.
To address this short recall interval, 
we first investigate the augmentation model with the replay method to ensure a sufficient recall interval in~\cref{sec:group_replay}.
Then, \cref{sec:group_ssl} introduces {\ourssl} learning model to extensively train multiple samples of the same instance in {\ourssl}.
\subsection{Replay with Augmentation}
\label{sec:group_replay}
Our view-batch model employs a replaying method with the augmentation to adjust the recall interval of retraining the same samples.
In the conventional learning scheduler, a sample-batch is replayed in the training process of neural networks. 
It means a batch for training networks consists of multiple unique training samples.
Therefore, the recall interval of retraining is the same as the number of training samples.
However, the strategy on our learning scheduler alternates the replaying method at the {\ourssl} level.
We structure a {\ourssl} to have multiple views of a single sample. 
Thus, the recall interval of a single sample increases with the size of view-batch. %
The concept of this view-batch model for the replay method is shown in~\cref{fig:f_learning_strategy}. 
It shows that when learning four samples in sequence, the time-space of re-training is four, whereas when learning $V$ views of a single sample, the time-space is extended by $V$ times.

Formally, we define samples as $I$, sample-batch as $\mathcal{B}^{I}=\{I_i\}_{i=1}^{B}$, the number of batches in a epoch as $T$, and learning scheduler as $\mathcal{A}$.
In the conventional learning algorithm, we replay each sample-batch to establish the scheduler.
\begin{equation}
    \mathcal{A}_\text{conventional} = [\underbrace{\mathcal{B}_1^I, \cdots, \mathcal{B}_{T}^I,}_{\text{recall}\;\text{interval}=B \times T} \mathcal{B}_1^I, \cdots, \mathcal{B}_{T}^I].
    \label{eq:batch_org}
\end{equation} 
Therefore, the recall interval of retraining is the same as the number of training samples $B \times T$ in an epoch.
However, to extend the recall interval, we replace the replaying unit from the sample-batch to the {\ourssl} $\mathcal{B}^\mathcal{V}=\{\mathcal{V}_i\}_{i=1}^{B}$.
We construct a {\ourssl} by including multiple views of the same sample $\mathcal{V}_i=\{I_i\}_{i=1}^{V}$, where $V$ denotes the number of views consisting of different augmentations of a single image. 
By modifying the replaying unit of~\cref{eq:batch_org} with a constructed view-batch, we define our new learning scheduler as follows: 
\begin{equation}
    \mathcal{A}_\text{ours} = [
    \underbrace{\mathcal{B}_1^\mathcal{V}, \cdots, \mathcal{B}_{T}^\mathcal{V},}_{\text{recall}\;\text{interval}=B \times T \times V} 
    \mathcal{B}_1^\mathcal{V}, \cdots, \mathcal{B}_{T}^\mathcal{V}
    ],
    \label{eq:batch_new}
\end{equation}
Based on this rescheduling, the recall interval is increased by $V$ times.
While we process $V$ views at a time, we decrease the total learning epoch by $V$, ensuring that the number of total samples that networks have seen during training is the same as the baseline method.
As a result, networks repeatedly forget and retrain samples during extended recall intervals, thereby improving long-term memory retention.
This learning of multiple views of a single sample adjusts the time-space while networks learn the same sample repeatedly.
To promote extensive learning, we present the self-supervised learning in the following.

\begin{algorithm}[t]
\caption{Pseudocode for our view-batch model.}
\label{alg:interleaved_block_learning}
\begin{algorithmic}[1]
\Require{dataset $\mathcal{D}^\text{train}$, parameters $\theta$, learning rate $\gamma$, batch size $B$, view-batch size $V$
}
\State Initialize network parameters $\theta$
\State Update $B \gets B / V$
\For{number of training iterations}
\State $\mathcal{B}^{I} = \{I_i\}_{i=1}^{B} \text{with } I_i \stackrel{\text{i.i.d.}}{\sim} D^\text{train}$
\For{$i=1, \cdots, B$}
\State $\mathcal{V}_i \gets \{I_i\}_{j=1}^{V}$
\Comment{replay of Equation (\textcolor{BrickRed}{2})}
\State $\mathcal{V}_{i,1} \gets A_w(\mathcal{V}_{i,1})$ 
\State $\mathcal{V}_{i,2:} \gets A_s(\mathcal{V}_{i,2:})$ 
\EndFor
\State $\mathcal{B}^{\mathcal{V}} \gets \{\mathcal{V}_i\}_{i=1}^{B}$
\State Compute $l_\text{sup} \gets L_\text{sup}(\theta, \mathcal{B}^{\mathcal{V}})$
\State Compute $l_\text{ssl} \gets L_\text{ssl}(\theta, \mathcal{B}^{\mathcal{V}})$
\Comment{SSL of Equation (\textcolor{BrickRed}{3})}
\State Update $\theta \gets \theta - \gamma \nabla_{\theta}(l_\text{sup} + l_\text{ssl})$ 
\EndFor
\end{algorithmic}
\end{algorithm}

\subsection{Self-supervised Learning}
\label{sec:group_ssl}
We present a self-supervised learning model with our VBM to extensively learn a sample at a time.
It is well-known that self-supervision generalizes the networks rather than acquiring task-specific knowledge~\citep{caron2021emerging}.
Further, task-agnostic knowledge from self-supervision is more robust to catastrophic forgetting of continual learning~\citep{cha2021co2l}.
For these reasons, we propose the SSL approach to effectively learn {\ourssl} in a \textit{(i) simple} and an \textit{(ii) efficient} manner.
For \textit{(i) simplicity}, unlike previous SSL methods~\citep{chen2020simple,grill2020bootstrap,caron2021emerging}, which require changes, such as architecture design~\citep{cha2021co2l} or teacher networks~\citep{cheng2024must}, we only modify an objective function to learn common knowledge within a {\ourssl} of same samples.
To satisfy \textit{(ii) efficiency}, our method does not use an extra training phase or longer training epochs than supervised learning, ensuring the minimal training cost as analyzed in the supplementary material.
Thus, the proposed method could be used in most continual learning methods without increasing computing costs as a drop-in replacement approach.

To implement our approach, we employ a one-to-many divergence loss to learn self-supervision within a view-batch $\{\mathcal{V}_i\}_{i=1}^{B}$. %
Our one-to-many divergence loss minimizes the average divergence between one weak-augmented view $A_{w}(\mathcal{V}_{i,1})$ and the remaining strong-augmented views $A_{s}(\mathcal{V}_{i,2:})$ of a view-batch at the logit level.
By aggregating different augmented views closer, networks learn common characteristics of objects.
Specifically, leveraging KL divergence $D_{\mathrm{KL}}$ between different augmentations, we define one-to-many-based self-supervised loss as 
\begin{equation}
    L_{\text{ssl}}(f_{\theta}, \mathcal{B}^\mathcal{V}) = 
    \frac{1}{B \cdot (V-1)}
    \sum_{i=1}^{B}
    \sum_{j=2}^{V}
    D_{\mathrm{KL}}(
    p_i^{1}
    ||
    p_{i}^{j}
    ),
    \label{eq:cgr_loss}
\end{equation}
where $p_i^{1}$ and $p_{i}^{j}$ represent network's prediction for weak augmented samples $\sigma(f_{\theta}(A_{w}(\mathcal{V}_{i,1})))$ and strong augmented samples $\sigma(f_{\theta}(A_{s}(\mathcal{V}_{i,j})))$, $f_\theta$ is neural networks, and $\sigma$ stand for the softmax. 
Finally, as shown in~\cref{alg:interleaved_block_learning}, we define our final objective function as
\begin{equation}
    \textrm{min}_{f_\theta} L_\text{sup}(f_\theta, \mathcal{B}^\mathcal{V}) + L_\text{ssl}(f_\theta, \mathcal{B}^\mathcal{V}),
\end{equation}
where $L_\text{sup}(f_\theta, \mathcal{B}^\mathcal{V}) = \frac{1}{B \cdot V}\sum_{i=1}^{B}\sum_{j=1}^{V}\mathcal{H}(y_i,p_i^j)$ of each view's prediction $p_i^j$ and its label $y_i$ using cross-entropy loss $\mathcal{H}$. We select widely used auto-augmentation~\citep{cubuk2018autoaugment} as a strong augmentation method, which slightly affects the performance of the baseline method as shown in~\cref{tab:factor_analysis_extended}, while using horizontal flip as weak augmentation.

\begin{figure}[t]
    \centering

    \hfill
    \hfill
    \begin{subfigure}[b]{.32\linewidth}
        \centering
        \includegraphics[width=\linewidth]{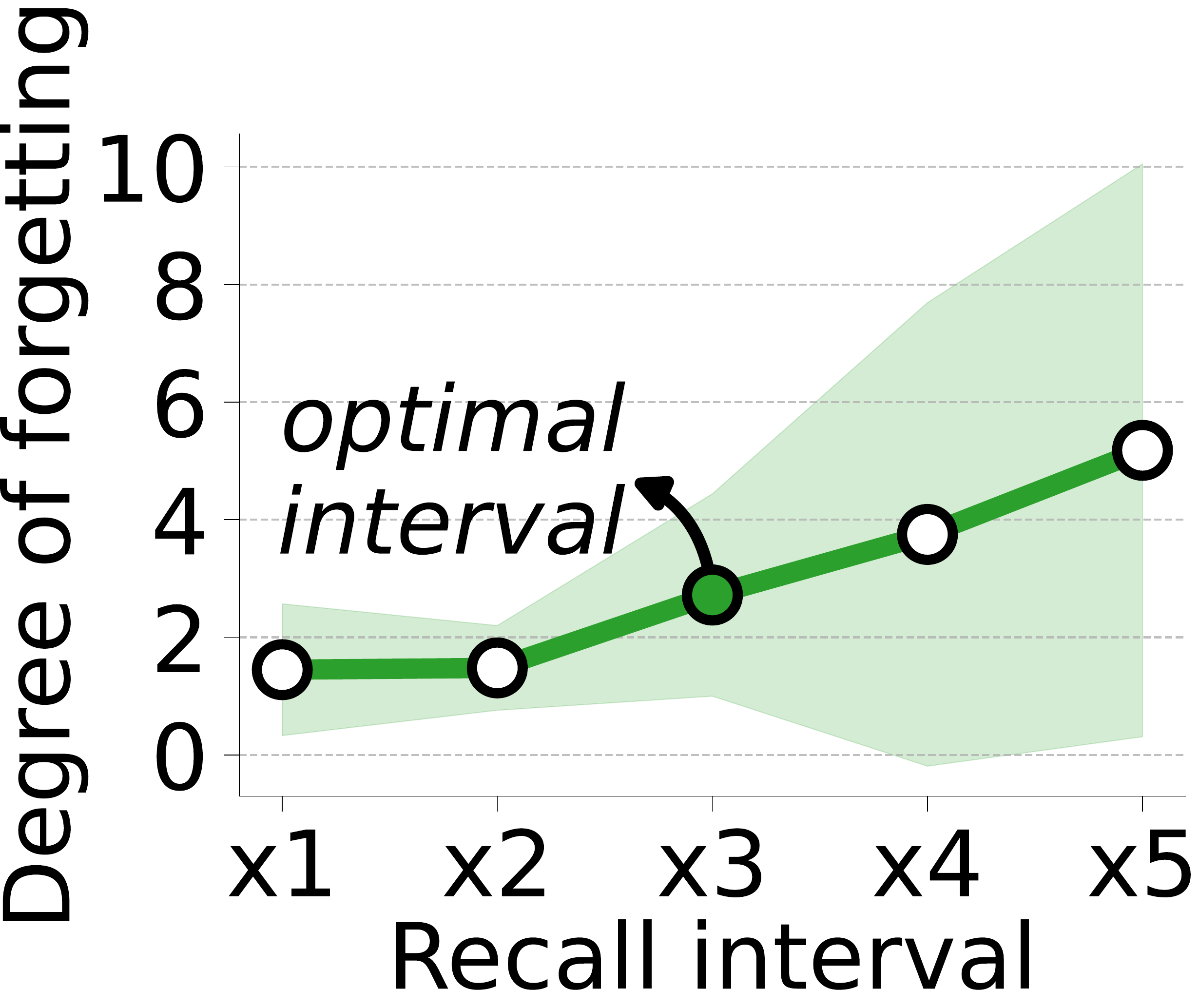}

        \caption{Degree of forgetting}
        \label{fig:p_forgetting}
    \end{subfigure}
    \hfill
    \begin{subfigure}[b]{.32\linewidth}
        \centering
        \includegraphics[width=\linewidth]{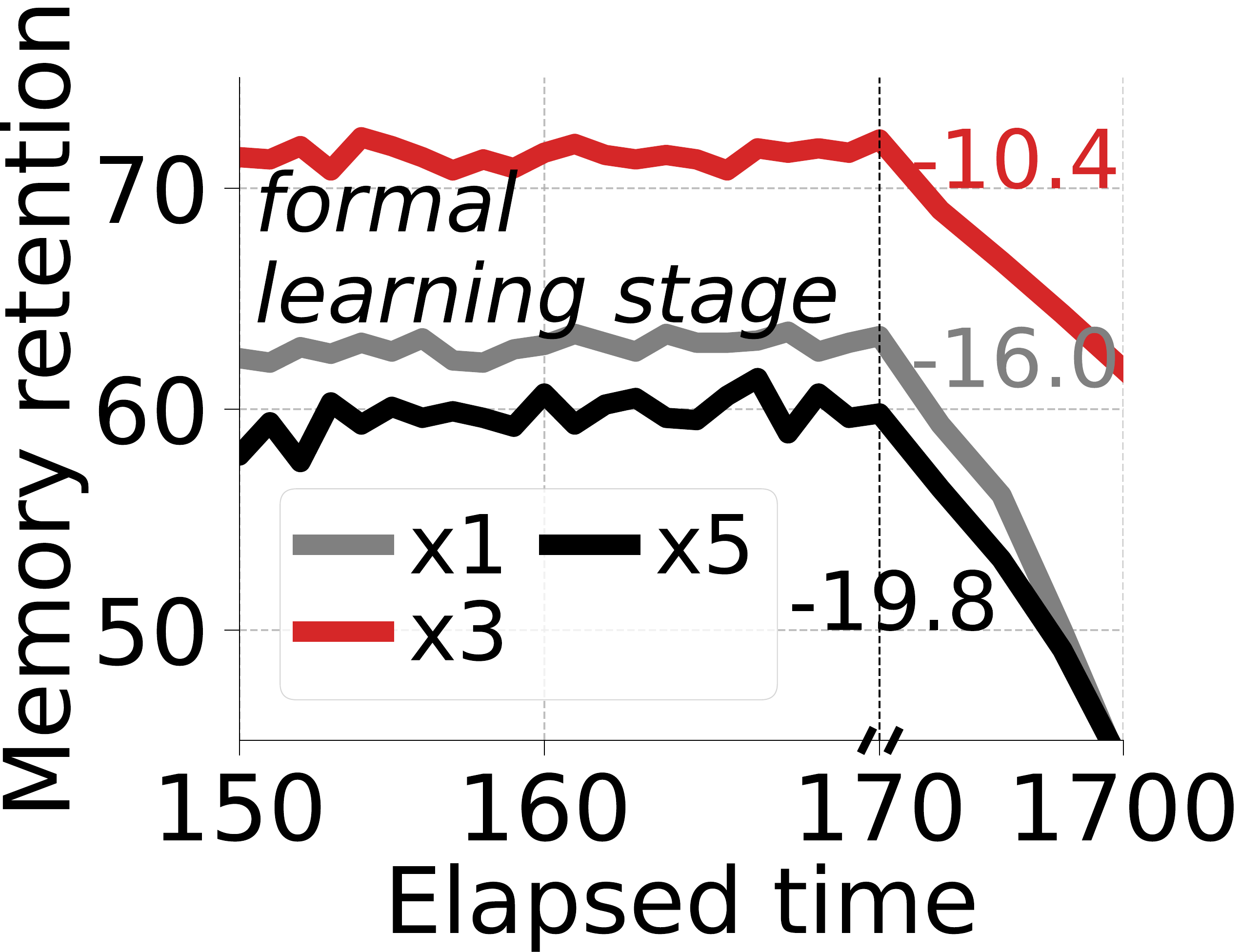}

        \caption{Forgetting curve}
        \label{fig:p_forgetting_curve}
    \end{subfigure}
    \hfill
    \begin{subfigure}[b]{.32\linewidth}
        \centering
        \includegraphics[width=\linewidth]{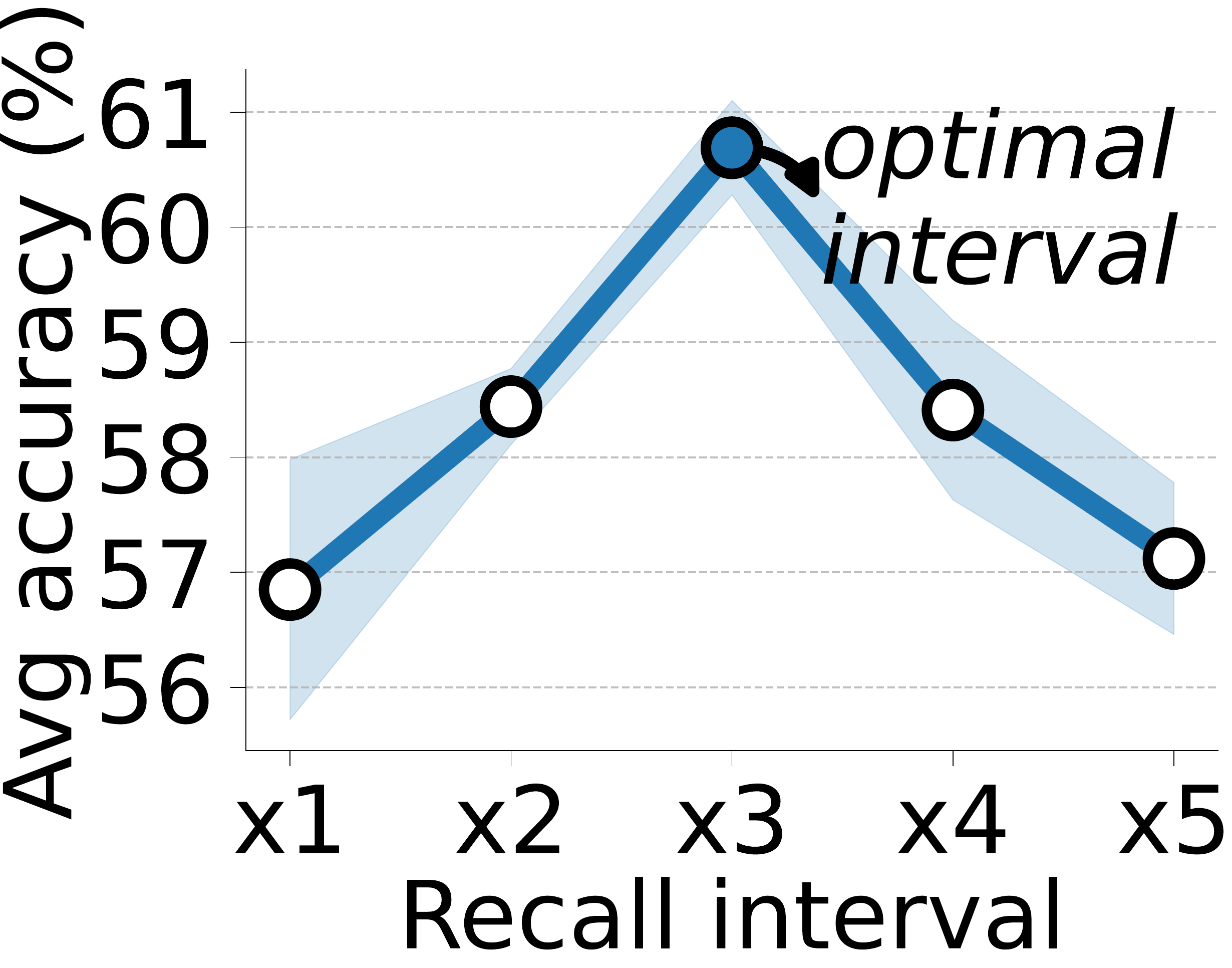}

        \caption{Accuracy}
        \label{fig:p_accuracy}
    \end{subfigure}
    \hfill
    \hfill
    \hfill
    
    \caption{
    \textbf{Empirical findings related to forgetting curve theory.}
    \textbf{(a)} We report the degree of forgetting for different recall intervals with a 95\% confidence interval denoted as shaded region. The degree of forgetting is measured as the performance degradation of each sample's classification accuracy between recall intervals. \textbf{(b)} We show the decay of memory retention as the learning progresses. This graph shows that when the formal learning stage is finished, memory retention decays over time for all three cases. 
    \textbf{(c)} We compare the network's classification accuracy of continual learning. It shows that \textbf{x3} achieves the best performance thanks to the slow memory retention decay. 
    }
    \label{fig:f_recall_interval}
\end{figure}

\begin{table*}[t]
    \centering
    \footnotesize
    \renewcommand{\arraystretch}{1} 
    \setlength{\tabcolsep}{6pt}

    \begin{tabular}{clcccccccc}
        \toprule
        
        \multirow{2}{*}{\textbf{Buffer}} &
        \multirow{2}{*}{\textbf{Method}} & 
        \multicolumn{4}{c}{\textbf{S-CIFAR-10}} &
        \multicolumn{4}{c}{\textbf{S-Tiny-ImageNet}} \\
        \cmidrule(lr){3-6}
        \cmidrule(lr){7-10}

        &
        &
        \textit{CIL} & 
        \textit{TIL} &
        \textit{Avg}* &
        $\Delta$ &
        \textit{CIL} & 
        \textit{TIL} &
        \textit{Avg}* &
        $\Delta$ \\
        \midrule

        - & Joint & 92.20\scriptsize{±0.15} & 98.31\scriptsize{±0.12} & 95.26 & ~ & 59.99\scriptsize{±0.19} & 82.04\scriptsize{±0.10} & 71.02 & ~ \\
        \midrule
        \multirow{2}{*}{0} & LwF & - & 61.98{±\scriptsize0.98} & 61.98 & ~ & - & 15.26{±\scriptsize0.71} & 15.26 & ~ \\
        ~ & \mycc VBM-LwF & \mycc - & \mycc 77.53\scriptsize{±3.06} & \mycc 77.53 & \mycc \textcolor{OliveGreen}{\textbf{+15.55}} & \mycc - & \mycc 51.21\scriptsize{±0.34} & \mycc 51.21 & \mycc \textcolor{OliveGreen}{\textbf{+35.95}} \\
        \midrule
        
        \multirow{7}{*}{200} & DER & 56.58\scriptsize{±2.08} & 89.32\scriptsize{±1.32} & 72.95 & ~ & 11.56\scriptsize{±0.58} & 41.67\scriptsize{±0.63} & 26.62 & ~ \\
        ~ & ER & 50.27\scriptsize{±0.70} & 91.66\scriptsize{±0.79} & 70.97 & ~ & 8.67\scriptsize{±0.23} & 38.03\scriptsize{±0.79} & 23.35 & ~ \\
        ~ & \mycc VBM-ER & \mycc \mycc 52.56\scriptsize{±1.59} & \mycc 93.58\scriptsize{±0.23} & \mycc 73.07 & \mycc \textcolor{OliveGreen}{\textbf{+2.10}} & \mycc 8.83\scriptsize{±0.13} & \mycc 44.03\scriptsize{±1.12} & \mycc 26.43 & \mycc \textcolor{OliveGreen}{\textbf{+3.08}} \\
        ~ & iCaRL & 64.11\scriptsize{±1.86} & 90.20\scriptsize{±0.88} & 77.16 & ~ & 19.09\scriptsize{±0.13} & 53.07\scriptsize{±0.27} & 36.08 & ~ \\
        ~ & \mycc VBM-iCaRL & \mycc 69.73\scriptsize{±0.87} & \mycc 92.76\scriptsize{±0.45} & \mycc 81.25 & \mycc \textcolor{OliveGreen}{\textbf{+4.09}} & \mycc 21.39\scriptsize{±0.18} & \mycc 56.31\scriptsize{±0.42} & \mycc 38.85 & \mycc \textcolor{OliveGreen}{\textbf{+2.77}} \\
        ~ & DER++ & 61.67\scriptsize{±1.02} & 90.61\scriptsize{±1.13} & 76.14 & ~ & 12.48\scriptsize{±1.60} & 41.19\scriptsize{±0.48} & 26.84 & ~ \\
        ~ & \mycc VBM-DER++ & \mycc 66.99\scriptsize{±2.98} & \mycc \mycc 94.30\scriptsize{±0.21} & \mycc 80.65 & \mycc \textcolor{OliveGreen}{\textbf{+4.51}} & \mycc 13.67\scriptsize{±0.41} & \mycc 44.36\scriptsize{±0.88} & \mycc 29.02 & \mycc \textcolor{OliveGreen}{\textbf{+2.18}} \\
        \midrule
        
        \multirow{7}{*}{500} & DER & 65.93\scriptsize{±5.38} & 92.52\scriptsize{±0.68} & 79.23 & ~ & 18.04\scriptsize{±1.30} & 52.90\scriptsize{±0.91} & 35.47 & ~ \\
        ~ & ER & 56.37\scriptsize{±0.96} & 94.15\scriptsize{±0.04} & 75.26 & ~ & 10.20\scriptsize{±0.37} & 49.13\scriptsize{±0.31} & 29.67 & ~ \\
        ~ & \mycc VBM-ER & \mycc 63.39\scriptsize{±2.77} & \mycc 94.20\scriptsize{±0.25} & \mycc 78.80 & \mycc \textcolor{OliveGreen}{\textbf{+3.54}} & \mycc 11.21\scriptsize{±0.30} & \mycc 52.01\scriptsize{±0.71} & \mycc 31.61 & \mycc \textcolor{OliveGreen}{\textbf{+1.94}} \\
        ~ & iCaRL & 62.22\scriptsize{±3.45} & 88.86\scriptsize{±3.23} & 75.54 & ~ & 23.93\scriptsize{±0.97} & 58.89\scriptsize{±0.85} & 41.41 & ~ \\
        ~ & \mycc VBM-iCaRL & \mycc 68.97\scriptsize{±0.38} & \mycc 93.10\scriptsize{±0.66} & \mycc 81.04 & \mycc \textcolor{OliveGreen}{\textbf{+5.50}} & \mycc 26.89\scriptsize{±0.54} & \mycc 62.61\scriptsize{±0.17} & \mycc 44.75 & \mycc \textcolor{OliveGreen}{\textbf{+3.34}} \\
        ~ & DER++ & 69.94\scriptsize{±2.04} & 93.49\scriptsize{±0.26} & 81.72 & ~ & 19.39\scriptsize{±1.09} & 51.71\scriptsize{±0.35} & 35.55 & ~ \\
        ~ & \mycc VBM-DER++ & \mycc 75.22\scriptsize{±0.81} & \mycc 94.81\scriptsize{±0.39} & \mycc 85.02 & \mycc \textcolor{OliveGreen}{\textbf{+3.30}} & \mycc 19.46\scriptsize{±0.46} & \mycc 49.80\scriptsize{±0.80} & \mycc 34.63& \mycc \textcolor{BrickRed}{\textbf{-0.92}} \\ 
        \midrule
        
        \multirow{7}{*}{5120} & DER & 80.79\scriptsize{±0.75} & 94.46\scriptsize{±0.30} & 87.63 & ~ & 32.92\scriptsize{±3.09} & 64.77\scriptsize{±2.14} & 48.85 & ~ \\
        ~ & ER & 74.23\scriptsize{±4.18} & 96.86\scriptsize{±0.28} & 85.55 & ~ & 28.07\scriptsize{±0.37} & 67.58\scriptsize{±0.30} & 47.83 & ~ \\
        ~ & \mycc VBM-ER & \mycc 79.29\scriptsize{±2.29} & \mycc 97.02\scriptsize{±0.13} & \mycc 88.16 & \mycc \textcolor{OliveGreen}{\textbf{+2.61}} & \mycc 33.07\scriptsize{±0.36} & \mycc 68.91\scriptsize{±0.50} & \mycc 50.99 & \mycc \textcolor{OliveGreen}{\textbf{+3.16}} \\
        ~ & iCaRL & 78.49\scriptsize{±0.38} & 94.84\scriptsize{±0.82} & 86.67 & ~ & 32.16\scriptsize{±0.24} & 66.75\scriptsize{±0.36} & 49.46 & ~ \\
        ~ & \mycc VBM-iCaRL & \mycc 79.40\scriptsize{±0.63} & \mycc 96.09\scriptsize{±0.32} & \mycc 87.75 & \mycc \textcolor{OliveGreen}{\textbf{+1.08}} & \mycc 34.40\scriptsize{±0.12} & \mycc 68.86\scriptsize{±0.26} & \mycc 51.63 & \mycc \textcolor{OliveGreen}{\textbf{+2.17}} \\
        ~ & DER++ & 85.42\scriptsize{±0.74} & 96.24\scriptsize{±0.19} & 90.83 & ~ & 34.06\scriptsize{±0.57} & 65.28\scriptsize{±1.20} & 49.67 & ~ \\
        ~ & \mycc VBM-DER++ & \mycc 86.62\scriptsize{±0.38} & \mycc 97.07\scriptsize{±0.07} & \mycc 91.85 & \mycc \textcolor{OliveGreen}{\textbf{+1.02}} & \mycc 35.14\scriptsize{±0.36} & \mycc 67.52\scriptsize{±0.48} & \mycc 51.33 & \mycc \textcolor{OliveGreen}{\textbf{+1.66}} \\
        \bottomrule
    \end{tabular}

    \caption{\textbf{Experimental results on CIL and TIL protocols.} 
    In this evaluation, we report the last top-1 accuracy (\%) on the S-CIFAR-10 and S-Tiny-ImageNet benchmarks with four different buffer sizes. 
    ResNet-18 backbone is used for each continual learning method in this evaluation.
    *\textit{Avg} is the averaged accuracy of \textit{CIL} and \textit{TIL}.
    We reproduce baseline methods following \citet{buzzega2020dark}. 
    We run every method three times with different random seeds for a reliable result. Ours improves performance consistently in most results.
    The performance of Joint is the upper bound of this experiment, where all old training samples are used in every step.
    }
    \label{tab:task_comp}
\end{table*}

\begin{table}[t]
    \centering
    \footnotesize
    \renewcommand{\arraystretch}{1.0} 
    \setlength{\tabcolsep}{6pt}
    \begin{tabular}{lcccc}
        \toprule
        \textbf{Method} & \textbf{Avg} & \textbf{$\Delta$} & \textbf{Last} & \textbf{$\Delta$} \\
        \midrule
        Joint & - & ~ & 46.48\scriptsize{±3.62} & ~ \\
        \midrule
        AGEM & 24.88\scriptsize{±0.57} & ~ & 28.94\scriptsize{±0.67} & ~ \\
        AGEM-R & 24.72\scriptsize{±0.59} & ~ & 28.01\scriptsize{±0.93} & ~ \\
        FDR & 34.62\scriptsize{±0.33} & ~ & 34.46\scriptsize{±0.32} & ~ \\
        DER & 39.84\scriptsize{±0.15} & ~ & 40.46\scriptsize{±0.47} & ~ \\
        \midrule
        ER & 32.38\scriptsize{±2.88} & ~ & 32.02\scriptsize{±2.39} & ~ \\
        \mycc VBM-ER & \mycc 35.80\scriptsize{±0.55} & \mycc \textcolor{OliveGreen}{\textbf{+3.42}} & \mycc 34.72\scriptsize{±0.37} & \mycc \textcolor{OliveGreen}{\textbf{+2.70}} \\
        DER++ & 34.56\scriptsize{±3.42} & ~ & 34.75\scriptsize{±2.41} & ~ \\
        \mycc VBM-DER++ & \mycc 42.16\scriptsize{±0.20} & \mycc \textcolor{OliveGreen}{\textbf{+7.24}} & \mycc 41.81\scriptsize{±0.08} & \mycc \textcolor{OliveGreen}{\textbf{+7.06}} \\
        \bottomrule
    \end{tabular}
    \caption{\textbf{Experimental results on DIL protocol.} 
    This evaluation shows \textit{Avg} and \textit{Last} top-1 accuracy (\%) on the DomainNet benchmark. We utilize ResNet-18 as the backbone network in this evaluation. 
    }
    \label{tab:domain_comp}
    \vspace{-1.em}
\end{table}

\begin{table*}[t]
    \centering
    \footnotesize
    
    \renewcommand{\arraystretch}{1} 
    \setlength{\tabcolsep}{4pt}
    
    \begin{tabular}{llclclclclclc}
        \toprule
        \multirow{2}{*}{\textbf{Method}} & 
        \multicolumn{4}{c}{\textbf{5 Step}} &
        \multicolumn{4}{c}{\textbf{10 Step}} & 
        \multicolumn{4}{c}{\textbf{20 Step}} \\
        \cmidrule(lr){2-5}
        \cmidrule(lr){6-9}
        \cmidrule(lr){10-13}

        &
        \multicolumn{1}{l}{\textit{Avg}} &
        $\Delta$ &
        \multicolumn{1}{l}{\textit{Last}} & 
        $\Delta$ &
        \multicolumn{1}{l}{\textit{Avg}} &
        $\Delta$ &
        \multicolumn{1}{l}{\textit{Last}} & 
        $\Delta$ &
        \multicolumn{1}{l}{\textit{Avg}} &
        $\Delta$ &
        \multicolumn{1}{l}{\textit{Last}} 
        $\Delta$ \\
        \midrule
        
        Joint & 80.40 & ~ & - & ~ & 80.41 & ~ & - & ~ & 81.49 & ~ & - \\ 
        \midrule
        iCaRL & 71.14{\scriptsize ±0.34} & ~ & 59.71 & ~ & 65.27{\scriptsize ±1.02} & ~ & 50.74 & ~ & 61.20{\scriptsize ±0.83} & ~ & 43.75 \\ 
        UCIR & 62.77{\scriptsize ±0.82} & ~ & 47.31 & ~ & 58.66{\scriptsize ±0.71} & ~ & 43.39 & ~ & 58.17{\scriptsize ±0.30} & ~ & 40.63 \\ 
        BiC & 73.10{\scriptsize ±0.55} & ~ & 62.10 & ~ & 68.80{\scriptsize ±1.20} & ~ & 53.54 & ~ & 66.48{\scriptsize ±0.32} & ~ & 47.02 \\ 
        WA & 72.81{\scriptsize ±0.28} & ~ & 60.84 & ~ & 69.46{\scriptsize ±0.29} & ~ & 53.78 & ~ & 67.33{\scriptsize ±0.15} & ~ & 47.31 \\ 
        DyTox & - & ~ & - & ~ & 73.66{\scriptsize ±0.02} & ~ & 60.67 & ~ & 72.27{\scriptsize ±0.18} & ~ & 56.32 \\ 
        \midrule
        
        DER & 76.77{\scriptsize ±0.01} & ~ & 68.06{\scriptsize ±0.00} & ~ & 75.72{\scriptsize ±0.08} & ~ & 64.32{\scriptsize ±0.05} & ~ & 74.96{\scriptsize ±0.01} & ~ & 61.80{\scriptsize ±0.07} \\ 
        \rowcolor[gray]{.9}
        VBM-DER & 78.60{\scriptsize ±0.23} & {\scriptsize \textcolor{OliveGreen}{\textbf{+1.83}}} & 70.60{\scriptsize ±0.12} & {\scriptsize \textcolor{OliveGreen}{\textbf{+2.54}}} & 78.12{\scriptsize ±0.07} & {\scriptsize \textcolor{OliveGreen}{\textbf{+2.40}}} & 67.04{\scriptsize ±0.11} & {\scriptsize \textcolor{OliveGreen}{\textbf{+2.72}}} & 76.95{\scriptsize ±0.02} & {\scriptsize \textcolor{OliveGreen}{\textbf{+1.99}}} & 64.29{\scriptsize ±0.04} & {\scriptsize \textcolor{OliveGreen}{\textbf{+2.49}}} \\ 
        TCIL & 77.33{\scriptsize ±0.08} & ~ & 69.48{\scriptsize ±0.14} & ~ & 76.33{\scriptsize ±0.15} & ~ & 65.66{\scriptsize ±0.02} & ~ & 74.32{\scriptsize ±0.01} & ~ & 62.54{\scriptsize ±0.02} \\ 
        \rowcolor[gray]{.9}
        VBM-TCIL & 79.23{\scriptsize ±0.01} & {\scriptsize \textcolor{OliveGreen}{\textbf{+1.90}}} & 71.23{\scriptsize ±0.04} & {\scriptsize \textcolor{OliveGreen}{\textbf{+1.75}}} & 78.02{\scriptsize ±0.03} & {\scriptsize \textcolor{OliveGreen}{\textbf{+1.69}}} & 68.14{\scriptsize ±0.08} & {\scriptsize \textcolor{OliveGreen}{\textbf{+2.48}}} & 76.83{\scriptsize ±0.00} & {\scriptsize \textcolor{OliveGreen}{\textbf{+2.51}}} & 67.16{\scriptsize ±0.09} & {\scriptsize \textcolor{OliveGreen}{\textbf{+4.62}}} \\
        \bottomrule
    \end{tabular}
    \caption{\textbf{Experimental results on rehearsal-based CIL protocol.} In this evaluation, we provide \textit{Avg} and \textit{Last} top-1 accuracy (\%) on the S-CIFAR-100 benchmark with three different class incremental steps. 
    ResNet-18 backbone is utilized to evaluate each rehearsal-based method. This evaluation also shows that ours achieves consistent performance improvements.
    Results of DyTox are imported from its original work \citep{douillard2022dytox}. 
    We reproduce results of DER and TCIL using official implementation, while other results come from \citet{yan2021dynamically}.
    }
    \label{tab:step_comp}
\end{table*}

\begin{table}[t]
    \centering
    \scriptsize

    \renewcommand{\arraystretch}{1} 
    \setlength{\tabcolsep}{.5pt}
    
    \begin{tabular}{lcccccccc}

        \toprule
        \multirow{2}{*}{\textbf{Method}} & 
        \multicolumn{4}{c}{\textbf{5 Step}} &
        \multicolumn{4}{c}{\textbf{10 Step}} \\
        \cmidrule(lr){2-5}
        \cmidrule(lr){6-9}

        &
        \multicolumn{1}{c}{\textit{Avg}} &
        \multicolumn{1}{c}{$\Delta$} & 
        \multicolumn{1}{c}{\textit{Last}} & 
        \multicolumn{1}{c}{$\Delta$} & 
        \multicolumn{1}{c}{\textit{Avg}} &
        \multicolumn{1}{c}{$\Delta$} & 
        \multicolumn{1}{c}{\textit{Last}} &
        \multicolumn{1}{c}{$\Delta$}
        \\
        \midrule

        Joint & 80.40 &  & - & & 80.41 &  & - \\
        \midrule
        
        LwF-MC & 32.89 & & 11.42 & & 20.71 & & 6.36 & \\ 
        BiC & 54.74 & & 35.42 & & 45.08 & & 24.42 & \\ 
        LwM & 39.31 & & 16.70 & & 25.47 & & 8.57 & \\ 
        WA & 63.67 & & 49.36 & & 45.68 & & 26.14 & \\ 
        \midrule

        DER & 40.05{\scriptsize ±0.00} & & 17.11{\scriptsize ±0.00} & & 26.67{\scriptsize ±0.00} & & 8.71{\scriptsize ±0.00} & \\ 
        \rowcolor[gray]{.9}
        VBM-DER & 40.47{\scriptsize ±0.00} & {\scriptsize\textcolor{OliveGreen}{\textbf{+0.42}}} & 17.47{\scriptsize ±0.01} & {\scriptsize\textcolor{OliveGreen}{\textbf{+0.36}}} & 26.95{\scriptsize ±0.00} & {\scriptsize\textcolor{OliveGreen}{\textbf{+0.28}}} & 8.90{\scriptsize ±0.00} & {\scriptsize\textcolor{OliveGreen}{\textbf{+0.19}}} \\ 
        
        TCIL & 64.40{\scriptsize ±0.15} & & 52.37{\scriptsize ±0.08} & & 56.84{\scriptsize ±0.75} & & 40.31{\scriptsize ±1.04} & \\ 
        \rowcolor[gray]{.9}
        VBM-TCIL & 66.40{\scriptsize ±0.07} & {\scriptsize\textcolor{OliveGreen}{\textbf{+2.00}}} & 54.69{\scriptsize ±0.02} & {\scriptsize\textcolor{OliveGreen}{\textbf{+2.32}}} & 61.12{\scriptsize ±0.02} & {\scriptsize\textcolor{OliveGreen}{\textbf{+4.28}}} & 45.92{\scriptsize ±0.25} & {\scriptsize\textcolor{OliveGreen}{\textbf{+5.61}}} \\   
        \bottomrule
    \end{tabular}

    \caption{
    \textbf{Experimental results on non-rehearsal-based CIL protocol.} 
    For this evaluation, we measure \textit{Avg} and \textit{Last} top-1 accuracy (\%) on the S-CIFAR-100 benchmark with two different class incremental steps. 
    ResNet-18 backbone is applied to existing and our methods.
    We report original results from LwF-MC and LwM \citep{dhar2019learning}.
    We reproduce the results of DER and TCIL by ourselves using official implementation.
    Other results are reported from \citet{huang2023resolving}.
    }
    \label{tab:non_rehearsal_step_comp}
\end{table}

\subsection{Empirical Evidence}%
\label{sec:spacing_effect}
This subsection empirically confirms our assumption that optimal recall interval improves memory retention, and accordingly, the performance of the continual method is enhanced.
Specifically, the logical development of our assumption is that 1) if the network sufficiently forgets a sample at the optimal recall interval and then extensively retrains it, 2) memory retention is enhanced as training progresses, and therefore, 3) the accuracy of continual learning is also improved.
To evaluate our assumption, we measure 
1) the \textbf{degree of forgetting} between a retraining session for training samples, 2) the decay speed of the memory retention, and 3) the \textbf{average accuracy} at which networks correctly classify their class labels.
~\cref{fig:f_recall_interval} shows the results for the empirical evidence of our assumption.
As depicted in~\cref{fig:p_forgetting}, the degree of forgetting becomes larger as the recall interval increases.
It is obvious that the longer recall interval allows the network to forget a training sample before retraining it again.
\cref{fig:p_forgetting_curve} demonstrates the different levels of memory retention decay due to the recall intervals.
The experiment results show that memory decays slowest in the \textbf{x3} case, which has an interval three times longer than the baseline \textbf{x1} case.
Notably, our method achieves considerably higher performance at the end of the formal learning stage.
This result is attributed to the fact that multiple epochs are used in training networks, and the memory retention decay between epochs is slow for the \textbf{x3} case.  
Accordingly, \textbf{x3} with our replay method achieves improved accuracy compared to other recall interval cases, as shown in~\cref{fig:p_accuracy}.
We conduct more experiments to show empirical proofs on different datasets, backbone networks, and buffer sizes in ~\cref{fig:recall_interval_analysis}.
The definition of forgetting is given in the supplementary material.

\begin{table*}[t]
    \centering
    \footnotesize

    \renewcommand{\arraystretch}{0.95} 
    \setlength{\tabcolsep}{6pt}
    
    \begin{tabular}{lcccccccc}

        \toprule
        & 
        \multicolumn{4}{c}{\textbf{S-CIFAR-100}} &
        \multicolumn{4}{c}{\textbf{S-ImageNet-R}} \\
        \midrule
        \midrule

        \multirow{2}{*}{\textbf{Method}} &
        \multicolumn{2}{c}{\textbf{Top-1}} &
        \multicolumn{2}{c}{\textbf{Top-5}} &
        \multicolumn{2}{c}{\textbf{Top-1}} &
        \multicolumn{2}{c}{\textbf{Top-5}} \\
        \cmidrule(lr){2-3}
        \cmidrule(lr){4-5}
        \cmidrule(lr){6-7}
        \cmidrule(lr){8-9}

        &
        \multicolumn{1}{c}{\textit{Avg}} & 
        \multicolumn{1}{c}{\textit{Last}} &
        \multicolumn{1}{c}{\textit{Avg}} & 
        \multicolumn{1}{c}{\textit{Last}} &
        \multicolumn{1}{c}{\textit{Avg}} & 
        \multicolumn{1}{c}{\textit{Last}} &
        \multicolumn{1}{c}{\textit{Avg}} & 
        \multicolumn{1}{c}{\textit{Last}} \\
        \midrule

        Joint & - & 92.54\scriptsize{±0.14} & - & 99.24\scriptsize{±0.02} & - & 82.41\scriptsize{±0.02} & - & 93.27\scriptsize{±0.07} \\
        \midrule
        L2P & 90.42\scriptsize{±0.56} & 83.76\scriptsize{±0.42} & 98.89\scriptsize{±0.03} & 98.08\scriptsize{±0.15} & 76.76\scriptsize{±0.33} & 73.56\scriptsize{±0.42} & 89.39\scriptsize{±0.16} & 86.61\scriptsize{±0.03} \\
        DualPrompt & 90.17\scriptsize{±0.38} & 85.10\scriptsize{±0.23} & 98.90\scriptsize{±0.07} & 97.68\scriptsize{±0.12} & 73.31\scriptsize{±0.19} & 69.12\scriptsize{±0.18} & 89.19\scriptsize{±0.09} & 85.46\scriptsize{±0.04} \\
        CODA-Prompt & 91.27\scriptsize{±0.66} & 86.86\scriptsize{±0.58} & 99.29\scriptsize{±0.09} & 98.64\scriptsize{±0.07} & 81.89\scriptsize{±0.59} & 76.79\scriptsize{±0.17} & 93.30\scriptsize{±0.34} & 90.49\scriptsize{±0.15} \\
        
        S-iPrompt & 92.19\scriptsize{±0.40} & 88.43\scriptsize{±0.42} & 99.19\scriptsize{±0.05} & 98.63\scriptsize{±0.05} & 71.47\scriptsize{±0.14} & 68.77\scriptsize{±0.27} & 88.44\scriptsize{±0.19} & 84.91\scriptsize{±0.31} \\
        
        \midrule
        SLCA & 94.31\scriptsize{±1.12} & 91.57\scriptsize{±0.46} & 99.64\scriptsize{±0.04} & 99.26\scriptsize{±0.03} & 83.58\scriptsize{±0.65} & 78.81\scriptsize{±0.21} & 94.39\scriptsize{±0.26} & 92.02\scriptsize{±0.25} \\ 

        \rowcolor[gray]{.9}
        VBM-SLCA &
        94.52\scriptsize{±0.85} &
        91.78\scriptsize{±0.18} &
        99.64\scriptsize{±0.03} &
        99.29\scriptsize{±0.03} & 
        84.61\scriptsize{±0.59} &
        80.07\scriptsize{±0.12} & 
        94.86\scriptsize{±0.19} & 
        92.74\scriptsize{±0.14} \\ 
        \bottomrule
    \end{tabular}

    \caption{
    \textbf{Experimental results on pre-trained model-based CIL protocol.} 
    This evaluation presents \textit{Avg} and \textit{Last} top-1,5 accuracy (\%) on the S-CIFAR-100 and S-ImageNet-R benchmarks without using a memory buffer. 
    ViT-B/16 backbone is adopted for each pre-trained model-based method.
    We reproduce baseline methods using official implementation by ourselves, except for L2P. %
    }
    \label{tab:prompt_comp}
\end{table*}

\section{Experiments}
\label{sec:exp}
This section demonstrates the capabilities of the proposed view-batch model in various continual learning protocols and scenarios through extensive performance evaluation.
We perform experiments for three continual learning protocols, such as class, task, and domain incremental learning.
Further, we also assess our view-batch model in both rehearsal and non-rehearsal, as well as under the pre-training-based continual learning scenarios.

\subsection{Experimental Settings}
\label{sec:exp_setting}

\paragraph{Datasets.}
We evaluate our methods on widely used benchmarks such as S-CIFAR-10/100, S-Tiny-ImageNet, S-ImageNet-R, and DomainNet where the prefix of `S-' denotes the sequential data configuration.
We use the commonly adopted class ordering lists, following existing works of~\citet{yan2021dynamically} and \citet{huang2023resolving}.

\paragraph{Protocols.}
In our experiments, we utilize three continual learning protocols.
First, we make use of task incremental learning (TIL), which involves training neural networks on a series of tasks in a sequential manner while evaluating their performance on both newly acquired and previously learned tasks.
Specifically, the TIL protocol consists of the following detailed configurations: 1) a task is defined by a set of label classes, and 2) we evaluate networks knowing the task identity of test samples.
Second, we adopt class incremental learning (CIL), which is identical to TIL except that it evaluates test samples without knowing their task identity.
Third, we use domain incremental learning (DIL), which focuses on learning the new samples across multiple sequential domains rather than acquiring new classes.
In addition to these three protocols, we also include both rehearsal and non-rehearsal scenarios, divided by the use of a memory buffer during continual learning.
For the non-rehearsal scenario, we evaluate both scratch- and pre-training-based learning scenarios to showcase extensive experimental results.
For all these protocols and scenarios, we provide the mean and standard deviation of three runs for reliable experimental results.
The details of the experiment can be found in the supplementary material.

\paragraph{Baseline Methods.}
We apply our view-batch to recent baseline continual learning methods such as iCaRL~\citep{rebuffi2017icarl}, 
ER~\citep{rolnick2019experience}, DER~\citep{buzzega2020dark,yan2021dynamically}, TCIL~\citep{huang2023resolving}, and SLCA~\citep{zhang2023slca}.
We also include results from A-GEM~\citep{chaudhry2018efficient}, LwM~\citep{dhar2019learning}, UCIR~\citep{hou2019learning}, BiC~\citep{wu2019large}, WA~\citep{zhao2020maintaining}, DyTox~\citep{douillard2022dytox}, L2P~\citep{wang2022learning}, 
S-iPrompt~\citep{wang2022s}, DualPrompt~\citep{wang2022dualprompt}, and CODA-Prompt~\citep{smith2023coda}.
We follow the baseline's hyper-parameters for a fair comparison between ours and its respective baseline methods.
Specifically, we use ResNet and Vision Transformer (ViT) as backbone networks in traditional and prompt-based approaches.
For the traditional approach, we randomly initialize backbone weights at the beginning of every run.
On the other hand, for the prompt-based approach, we adopt a frozen backbone whose weights are pre-trained on a large-scale dataset following recent prompt-based CL studies~\citep{wang2022learning,zhang2023slca} %

\begin{figure}[t]
    \centering

    \hfill
    \hfill
    \begin{subfigure}[t]{0.45\linewidth}
        \centering        
        \includegraphics[width=\linewidth]{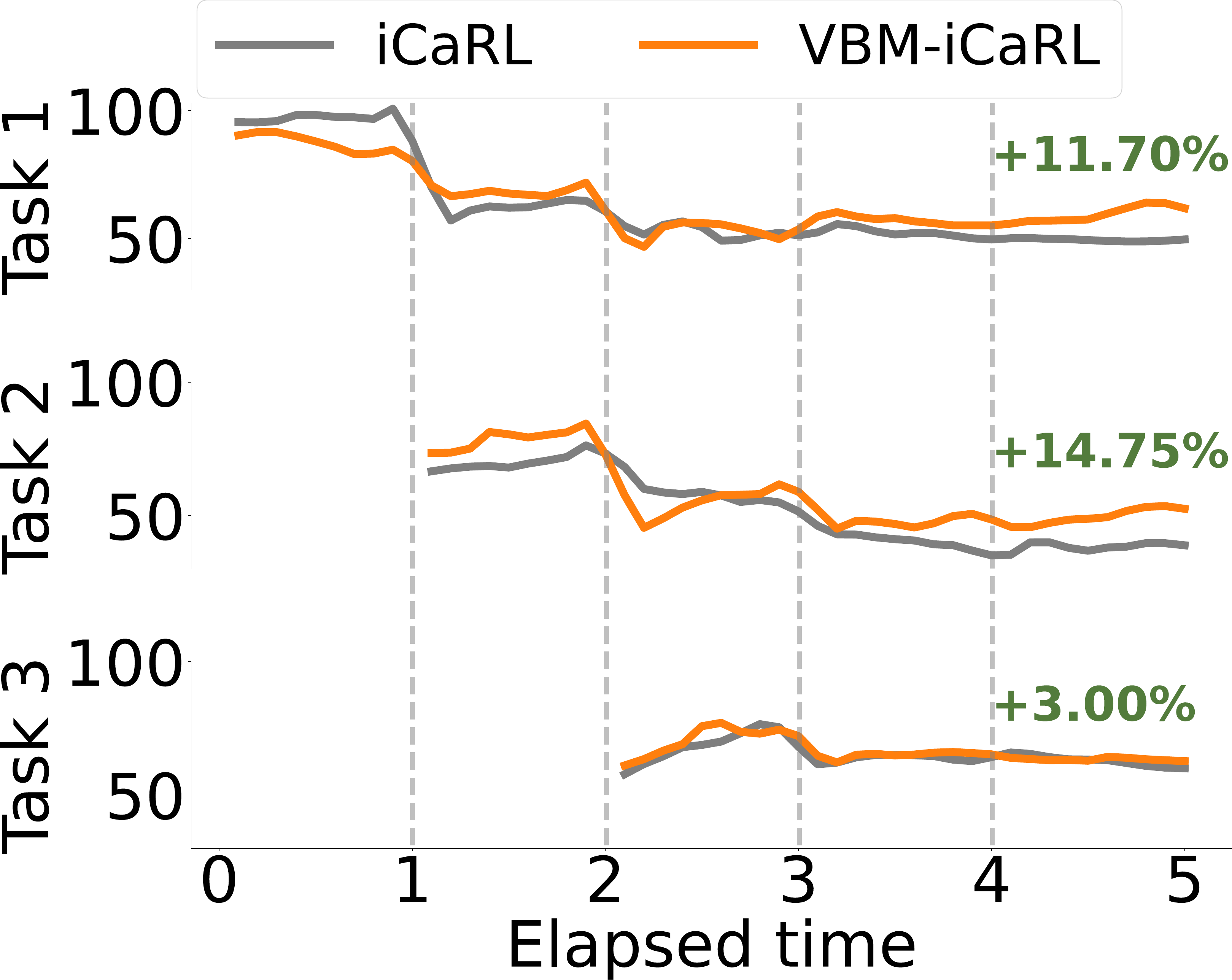}
        \caption{iCaRL}
        \label{fig:f_memory_retention_decay_icarl}
    \end{subfigure}
    \hfill
    \begin{subfigure}[t]{0.45\linewidth}
        \centering        
        \includegraphics[width=\linewidth]{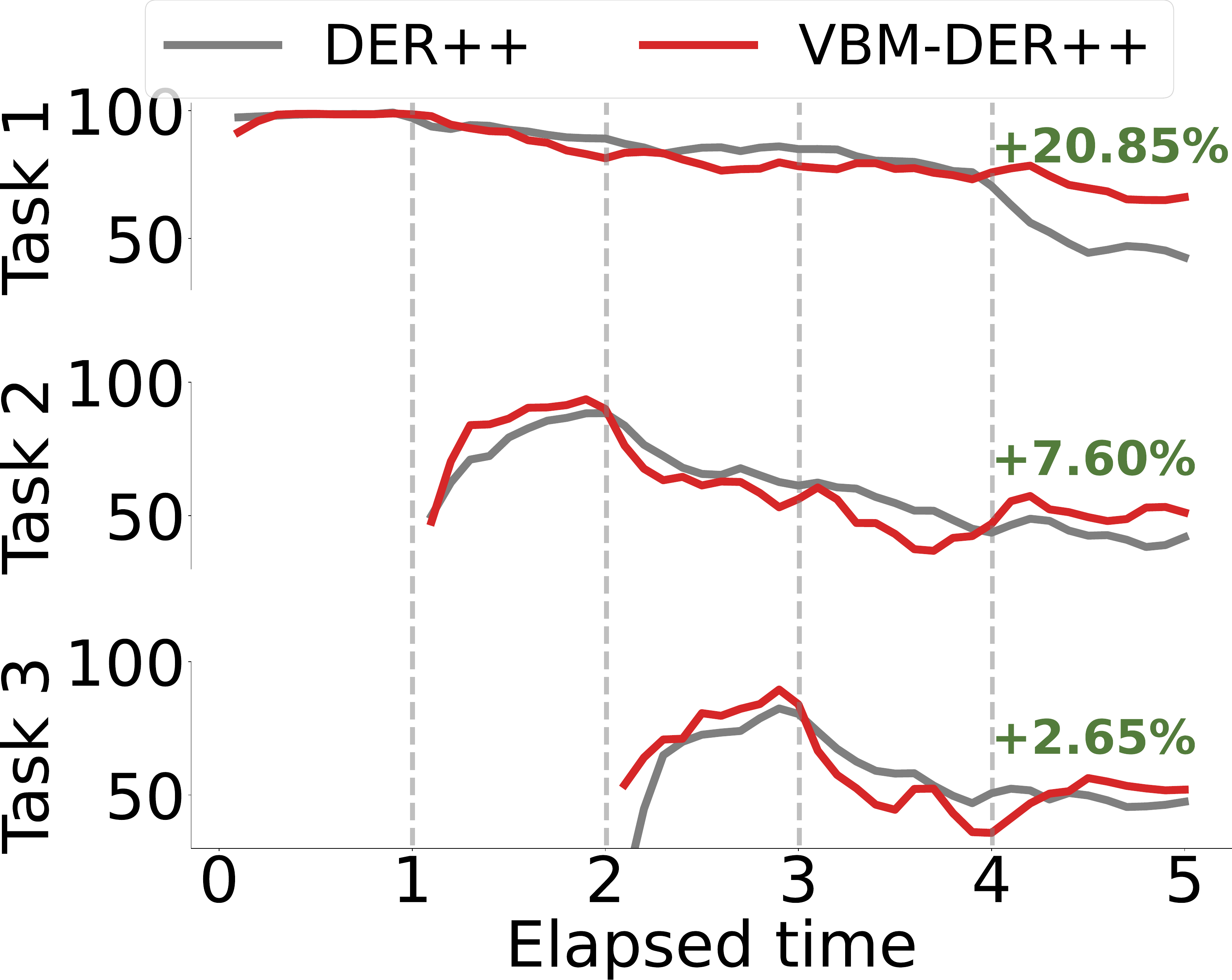}
        \caption{DER++}
        \label{fig:f_memory_retention_decay_derpp}
    \end{subfigure}
    \hfill
    \hfill
    \hfill
    
    \caption{
    \textbf{Experimental results on memory retention decay.}
    This analysis reports memory retention decay of the first three tasks on the S-CIFAR-10 dataset, comparing the proposed approach against the baseline methods.
    The \textcolor{OliveGreen}{\textbf{green}} numbers at the end of the last task are the accuracy gain of the proposed approach over the baselines.
    }
    \label{fig:f_memory_retention_decay}
\end{figure}

\begin{figure}[t]
    \centering
    \hfill\hfill
    \begin{subfigure}[]{.3\linewidth}
        \includegraphics[width=\linewidth]{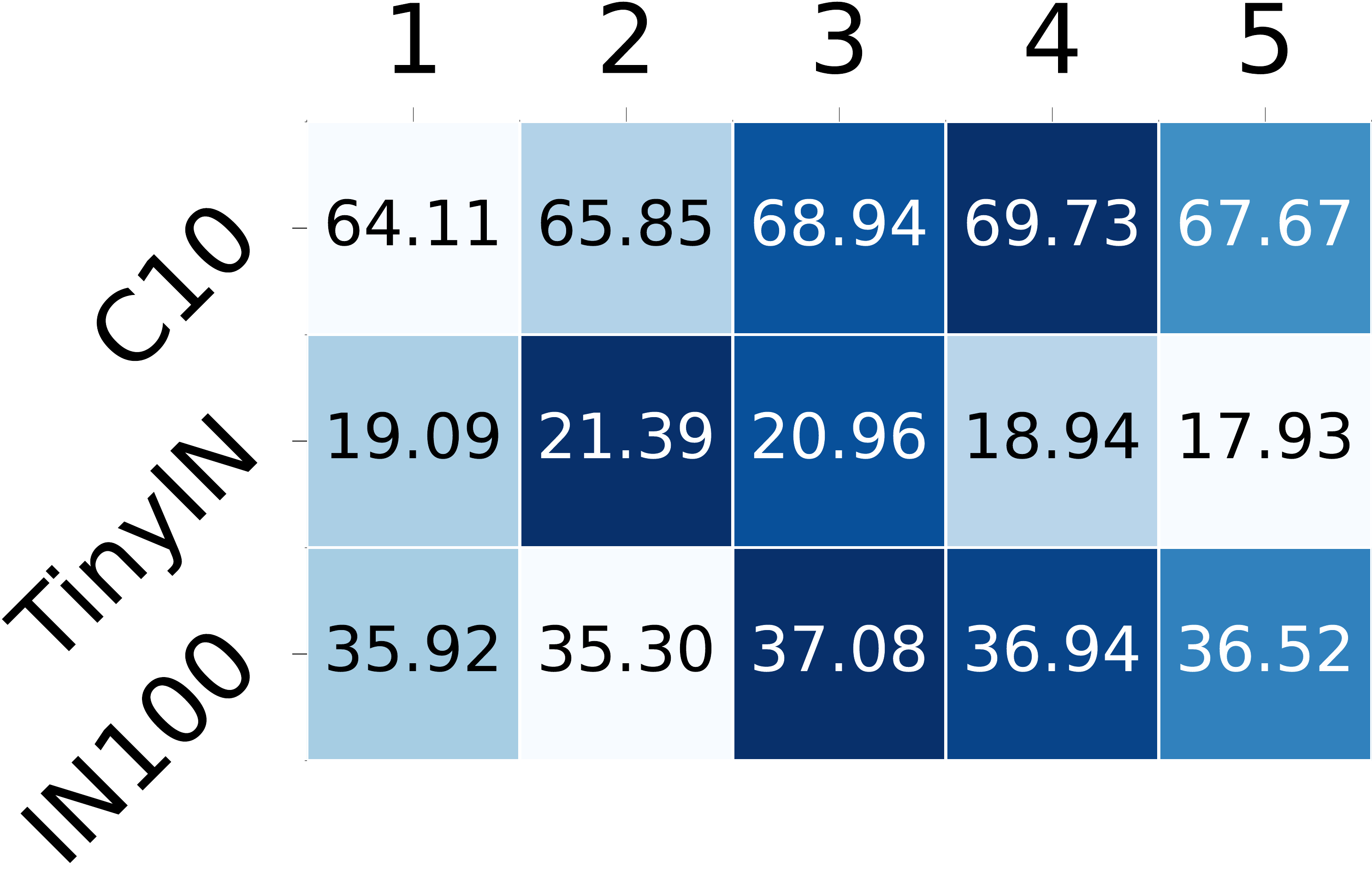}
        
        \includegraphics[width=\linewidth]{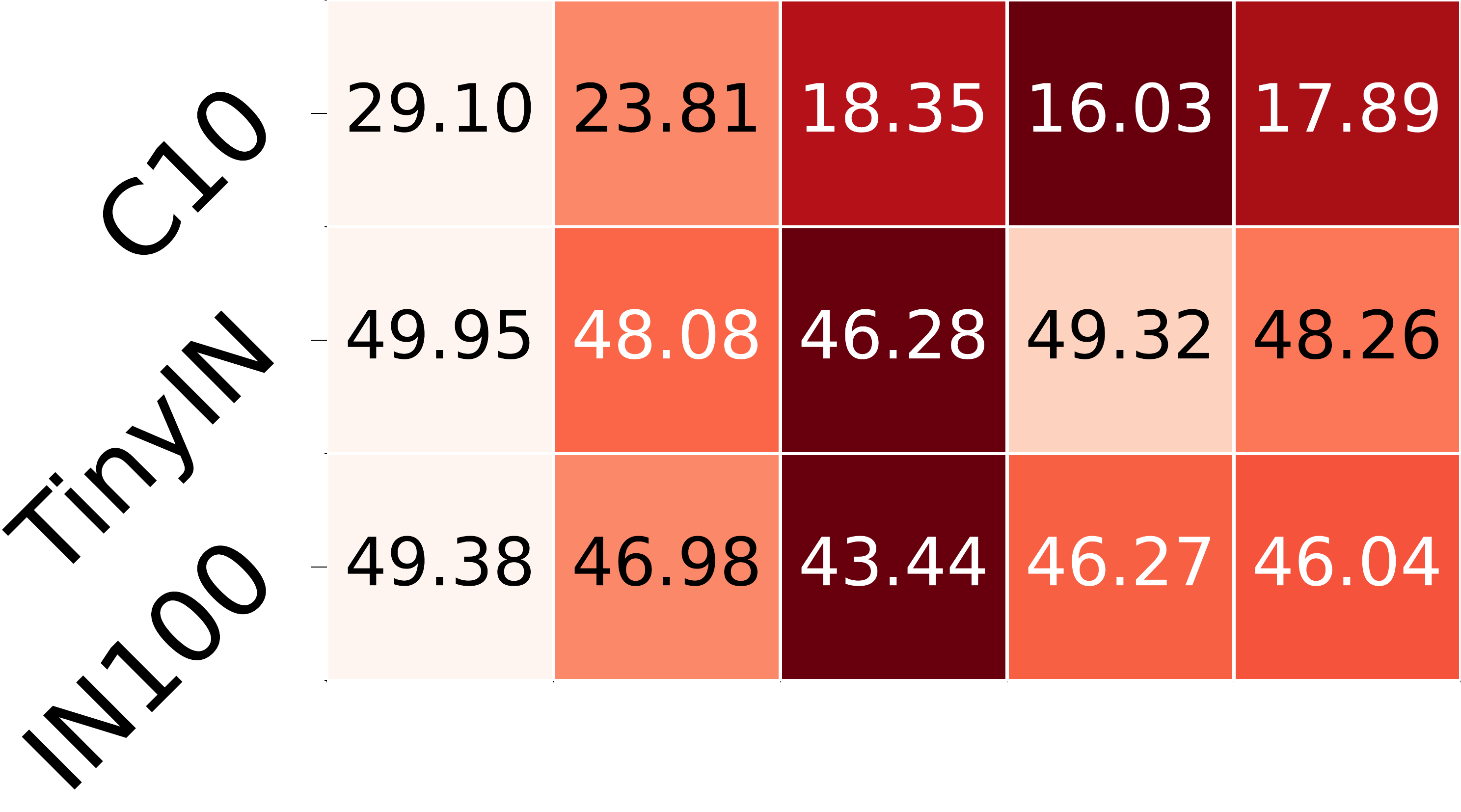}

        \vspace{-.25em}
        
        \caption{Dataset}
    \end{subfigure}
    \hfill
    \begin{subfigure}[]{.3\linewidth}
        \includegraphics[width=\linewidth]{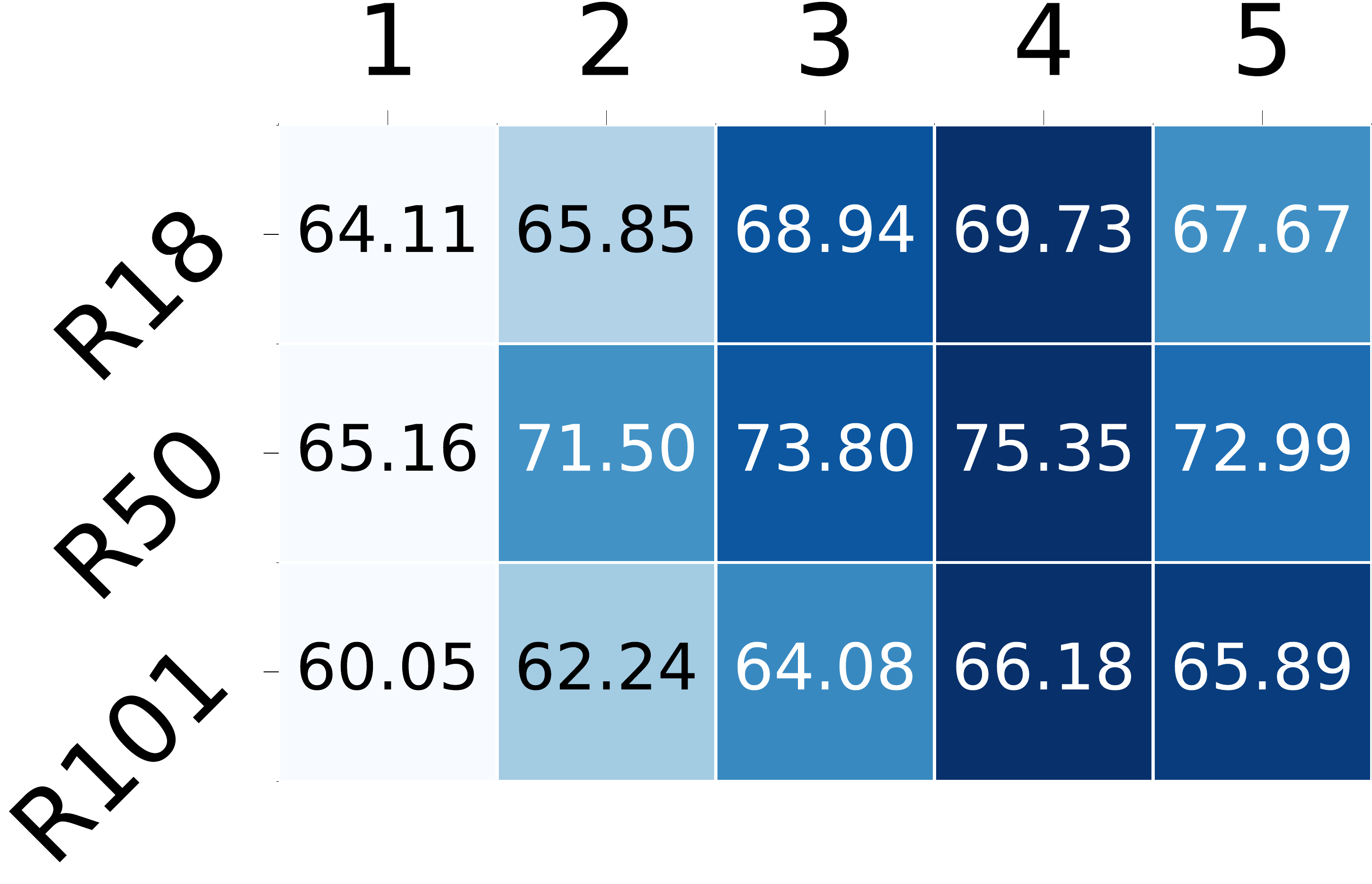}
        
        \includegraphics[width=\linewidth]{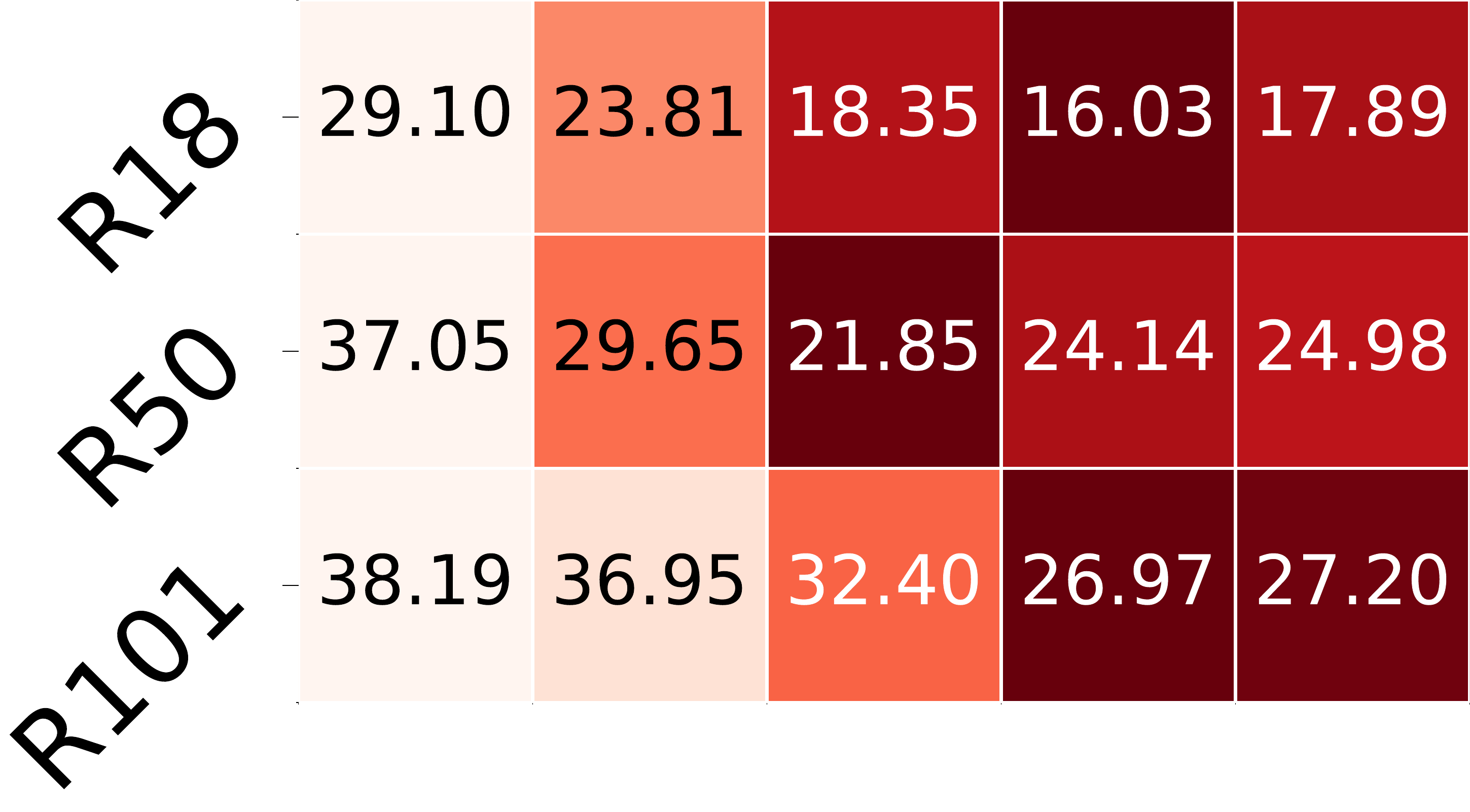}

        \vspace{-.25em}
        
        \caption{Network}
    \end{subfigure}
    \hfill
    \begin{subfigure}[]{.3\linewidth}
        \includegraphics[width=\linewidth]{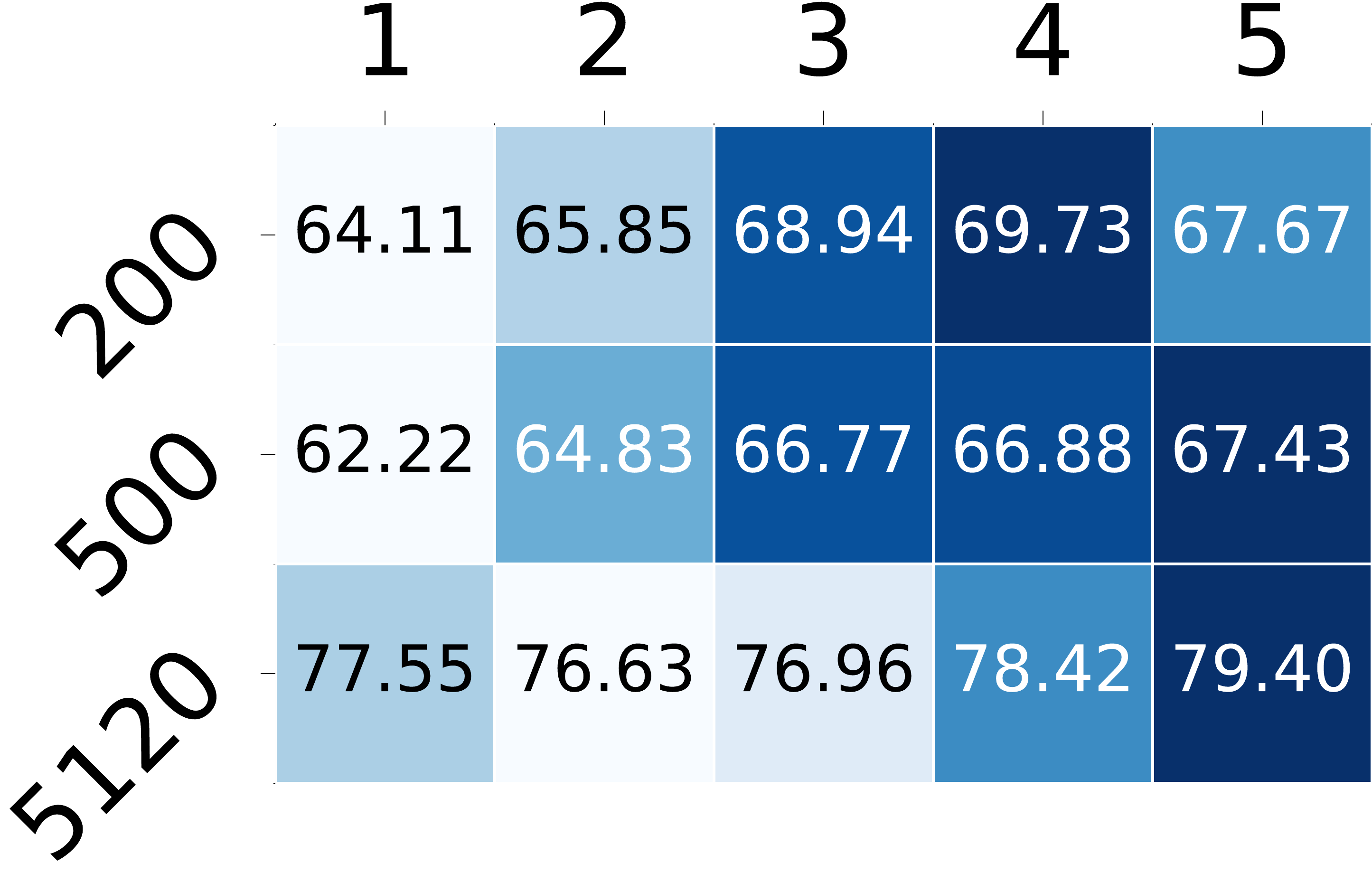}
        
        \includegraphics[width=\linewidth]{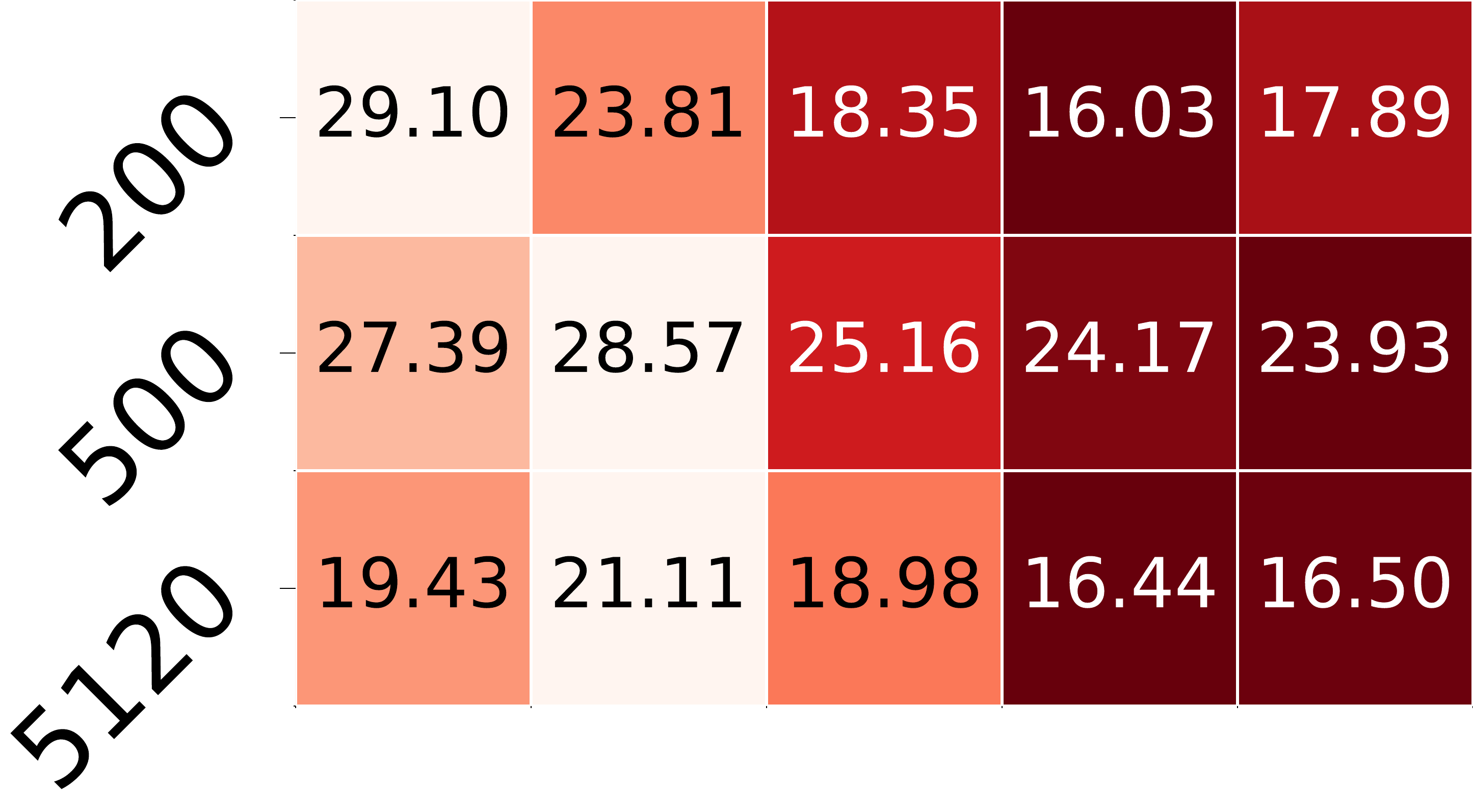}

        \vspace{-.25em}
        
        \caption{Buffer}
    \end{subfigure}
    \hfill\hfill\hfill

    \caption{\textbf{Experimental results on recall interval.} We measure the last top-1 accuracy (\%) (\textcolor{blue}{$\uparrow$}) in the top row and report forgetting (\textcolor{red}{$\downarrow$}) at the bottom row, varying recall intervals with different learning factors. Dark colors mean better. We use the ResNet-18 backbone.}
    \label{fig:recall_interval_analysis}
\end{figure}

\subsection{Experimental Results}
\paragraph{Continual Learning under Three Protocols.}
\label{sec:exp_scl}
We evaluate our method in the three continual learning protocols such as CIL, TIL, and DIL.
For TIL and CIL, we use different memory buffer sizes: 0, 200, 500, and 5,120.
For DIL, we utilize 50 samples per class, resulting in 17,250 samples in a buffer.
We apply our view-batch model to LwF for the non-rehearsal memory buffer (\ie 0 size) and ER, DER++, and iCaRL for the other buffer sizes.
\cref{tab:task_comp,tab:domain_comp} show that our method improves the performance of continual learning methods for all protocols.
Moreover, we demonstrate that baseline methods employing our view-batch are comparable to other continual learning methods.

\begin{table}
    \centering
    \footnotesize
    \renewcommand{\arraystretch}{0.95} 
    \setlength{\tabcolsep}{4pt}
    \begin{tabular}{cccccccc}
        \toprule

        \textbf{Method} &
        \textbf{Replay} &
        \specialcell{\textbf{Strong}\\\textbf{augment}} &
        \textbf{SSL} &
        \textbf{CIL} &
        \textbf{TIL} &
        \textbf{Avg} &
        \textbf{$\Delta$}
        \\
        \midrule
        
        \multirow{5}{*}{iCaRL} & - & - & - & 64.11 & 90.20 & 77.16 & \\
        & - & \boldcheckmark & - &  62.04 & 90.17 & 76.11 & \textcolor{BrickRed}{\textbf{-1.05}} \\
        & \boldcheckmark & - & - & 67.51 & 91.87 & 79.69 & \textcolor{OliveGreen}{\textbf{+2.53}}  \\
        & \checkmark & \boldcheckmark & - & 66.78 & 91.21 & 78.99	 & \textcolor{OliveGreen}{\textbf{+1.83}}  \\
        & \mycc \checkmark & \mycc\checkmark & \mycc\boldcheckmark & \mycc69.73 & \mycc92.76 & \mycc81.25 & \mycc\textcolor{OliveGreen}{\textbf{+4.09}} \\
        \midrule

        \multirow{5}{*}{DER++} & - & - & - & 61.67 & 90.61 & 76.14 & \\
        & - & \boldcheckmark & - &  62.66 & 91.14 & 76.90 & \textcolor{OliveGreen}{\textbf{+0.76}}  \\
        & \boldcheckmark & - & - & 64.75 & 92.79 & 78.77 & \textcolor{OliveGreen}{\textbf{+2.63}} \\
        & \checkmark & \boldcheckmark & - & 64.44 & 93.38 & 78.91 & \textcolor{OliveGreen}{\textbf{+2.77}} \\
        & \mycc\checkmark & \mycc\checkmark & \mycc\boldcheckmark & \mycc66.99 & \mycc94.30 & \mycc80.65 &  \mycc\textcolor{OliveGreen}{\textbf{+4.51}} \\
        \bottomrule
    \end{tabular}
    \caption{
    \textbf{Experimental results on factor analysis.}
    We demonstrate the last accuracy (\%) of CIL and TIL on S-CIFAR-10, varying our main components with different augmentation types. 
    We observe that the proposed components, such as replay and SSL, improve performance consistently and confirm that performance enhancement comes from them, not simply from strong augmentation.
    In the strong augmentation setting without view batch replay, we maintain the same ratio of weak to strong augmentations.} %
    \label{tab:factor_analysis_extended}
\end{table}

We conduct further in-depth experiments under the most widely used CIL protocol.
We consider three configurations with step sizes of 5, 10, and 20, which are the number of total incremental steps, as shown in ~\cref{tab:step_comp}. 
In addition, we take into account the use of a memory buffer in two scenarios, such as rehearsal and non-rehearsal in \cref{tab:non_rehearsal_step_comp}.
Both experimental results demonstrate that our view-batch model consistently improves the performance of continual learning methods under the various step sizes and memory buffer configurations.

\paragraph{Non-rehearsal Continual Learning.}
\label{sec:exp_pcl}
While the non-rehearsal scenario has the advantage of not using memory buffers, they inherently face the drawback of limited performance due to the inability to retrain networks using old data in the memory buffer.
To overcome this limitation, many recent studies have focused on non-rehearsal continual learning using a pre-trained model trained with a large-scale dataset.
Thus, we evaluate our view-batch model in a scenario where models are pre-trained, such as SCLA.
Experimental results in \cref{tab:prompt_comp} indicate that the proposed view-batch model is effective across all benchmarks for this scenario.

\begin{table}
    \centering
    \footnotesize
    \renewcommand{\arraystretch}{0.95} 
    \setlength{\tabcolsep}{5.5pt}

    \begin{tabular}{cccrrrr}
        \toprule

        \textbf{Method} & 
        \specialcell{\textbf{Current}\\\textbf{sample}} & 
        \specialcell{\textbf{Buffer}\\\textbf{sample}} & 
        \specialcell{\textbf{CIL}} &
        \specialcell{\textbf{TIL}} &
        \specialcell{\textbf{Avg}} &
        \specialcell{\textbf{$\Delta$}}
        \\
        \midrule
        
        \multirow{4}{*}{iCaRL} & - & - & 64.11 & 90.20 & 77.16 & - \\
        & \boldcheckmark & - & 64.64 & 94.12 & 79.38 & \textcolor{OliveGreen}{\textbf{+2.22}} \\
        & - & \boldcheckmark & 64.99 & 94.20 & 79.60 & \textcolor{OliveGreen}{\textbf{+2.44}}\\
        & \mycc\boldcheckmark & \mycc\boldcheckmark & \mycc 69.73 & \mycc 92.76 &\mycc 81.25 &\mycc \textcolor{OliveGreen}{\textbf{+4.09}} \\
        \midrule
        
        \multirow{4}{*}{DER++} & - & - & 61.67 & 90.61 & 76.14 & - \\
        & \boldcheckmark & - & 64.88 & 93.22 & 79.05 & \textcolor{OliveGreen}{\textbf{+2.91}} \\
        & - & \boldcheckmark & 62.69 & 94.15 & 78.42 & \textcolor{OliveGreen}{\textbf{+2.28}} \\
        & \mycc\boldcheckmark & \mycc\boldcheckmark & \mycc 66.99 & \mycc 94.30 &\mycc 80.65 &\mycc\textcolor{OliveGreen}{\textbf{+4.51}} \\
        
        \bottomrule
    \end{tabular}

    \caption{
    \textbf{Experimental results on sample type.}
    We measure the last accuracy (\%) of CIL and TIL on S-CIFAR-10, applying our method to different sample types. 
    We use the ResNet-18 as backbone.
    This analysis shows that our method works well with both sample types.
    }
    \label{tab:sample_type_analysis}
\end{table}

\section{Analysis}
\label{sec:analysis}
We present the analysis of the proposed view-batch model. %
First, we validate that the optimal recall interval mitigates catastrophic forgetting of early tasks.
\cref{fig:f_memory_retention_decay} visualize that the recall interval \textbf{x3} maintains long-term memory retention resolving catastrophic forgetting of initial tasks.
Also, \cref{fig:recall_interval_analysis} analyzes how different factors (\textit{i.e.}, dataset, network scales, memory buffer size) affect the optimal recall intervals in terms of last top-1 accuracy and forgetting metrics~\citep{chaudhry2018riemannian,chaudhry2018efficient,wang2022learning}. We find that the recall interval \textbf{x3} or \textbf{x4} improves the accuracy and resolves catastrophic forgetting in various scenarios. More experimental results of different methods and task types can be found in the supplementary material.
\cref{tab:factor_analysis_extended} shows that two main factors of the proposed method consistently improve their respective baselines, identifying that the performance improvement comes from the proposed components, not strong augmentation.
Lastly, in the rehearsal scenario, we have two different sample types, current- and buffer-based samples.
\cref{tab:sample_type_analysis} exhibit that ours performs well with both types.

\begin{figure}[t]
    \centering
    
    \hfill
    \hfill
    \hfill
    \hfill
    \includegraphics[width=.87\linewidth]{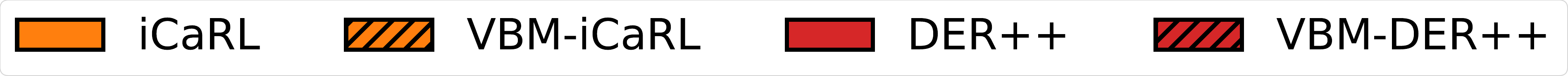}
    \hfill
    \hfill
    \hfill

    \vspace{.3em}

    \hfill
    \hfill
    \begin{subfigure}[t]{0.45\linewidth}
        \centering        
        \includegraphics[width=\textwidth]{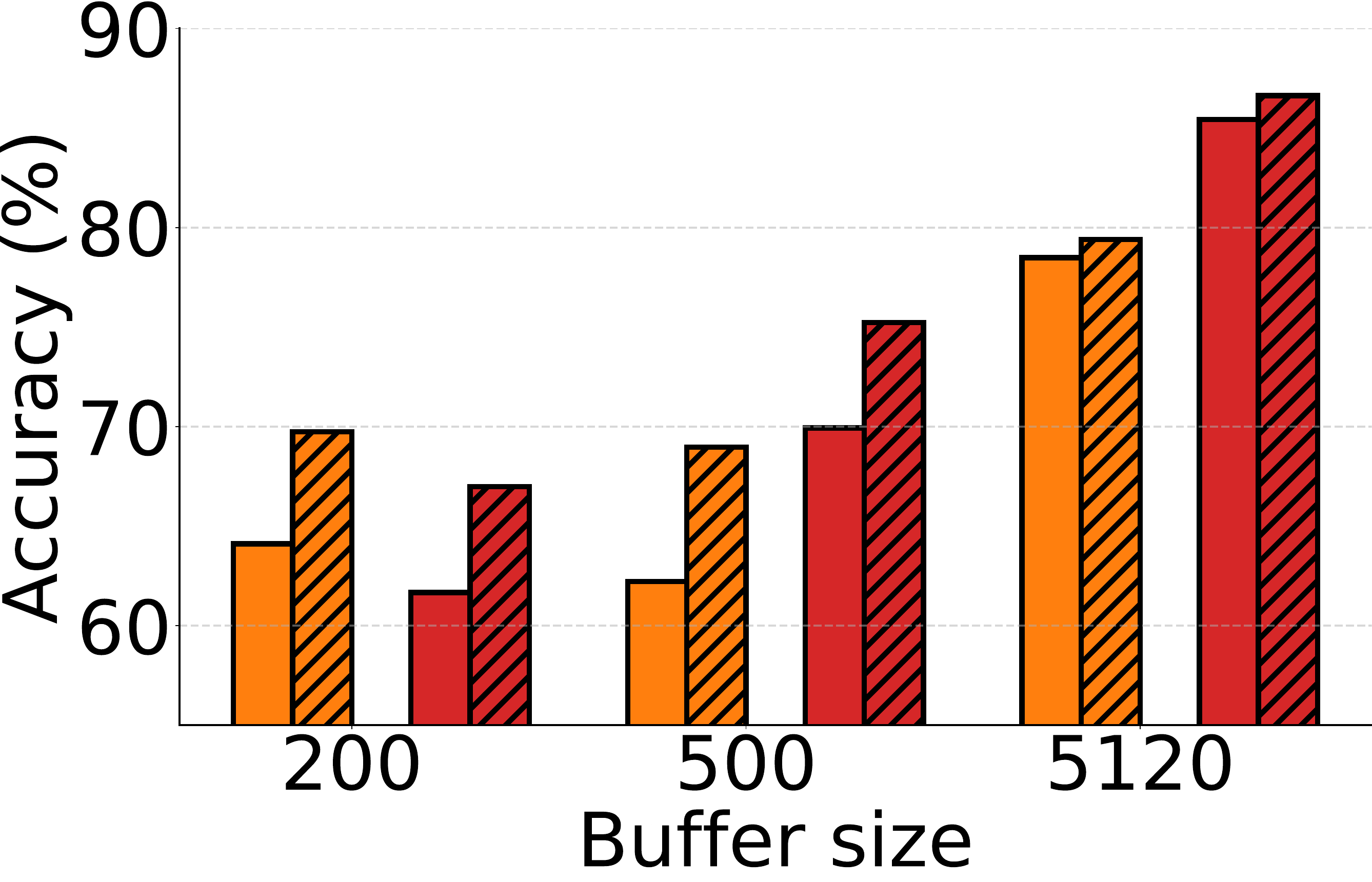}
        \caption{Accuracy ($\uparrow$)}
        \label{fig:f_last_acc}
    \end{subfigure}
    \hfill
    \begin{subfigure}[t]{0.45\linewidth}
        \centering
        \includegraphics[width=\textwidth]{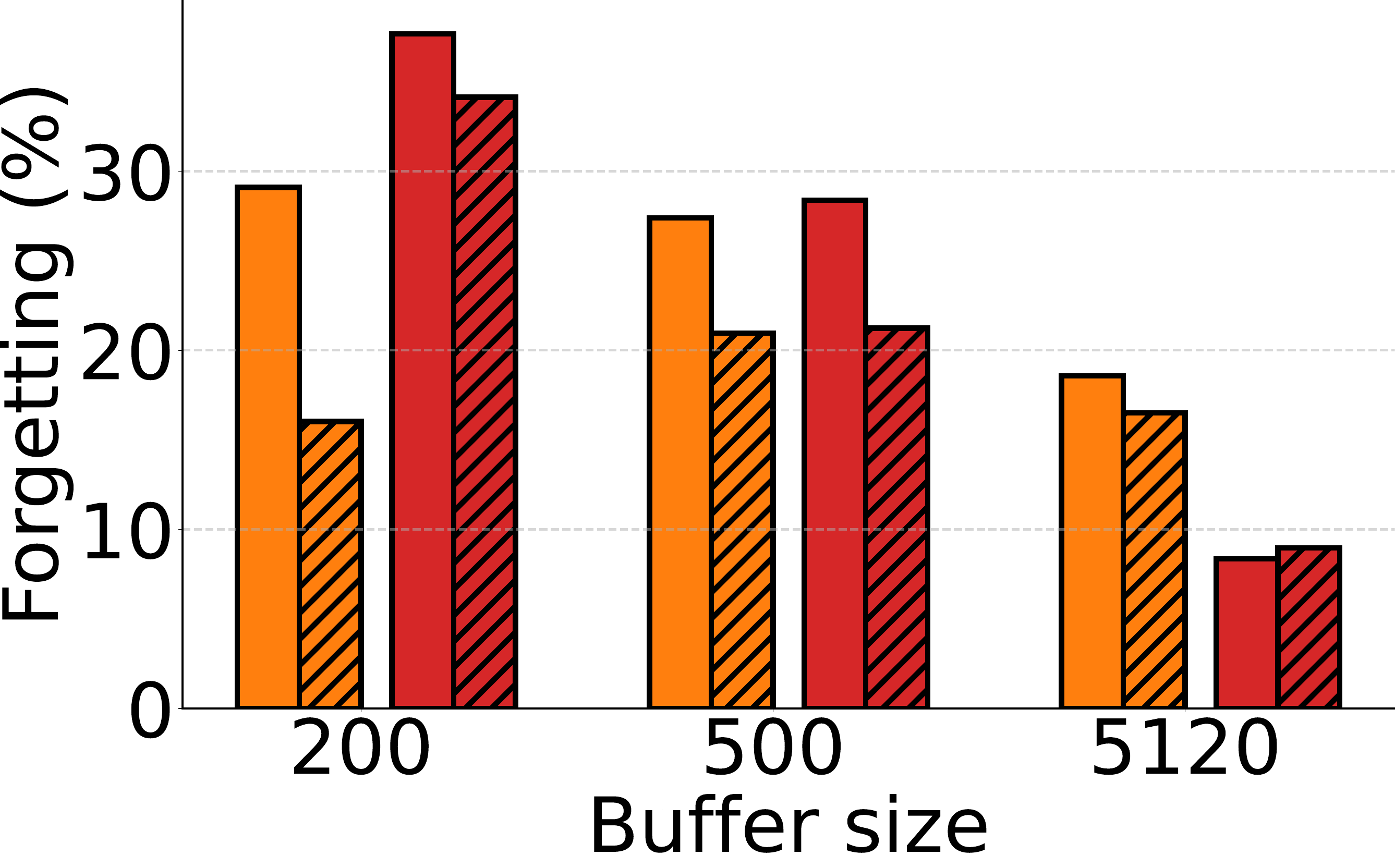}
        \caption{Forgetting ($\downarrow$)}
        \label{fig:f_forgetting}
    \end{subfigure}
    \hfill
    \hfill
    \hfill
    
    \caption{
    \textbf{Experimental results on accuracy and forgetting.}
    This analysis shows (a) the last top-1 accuracy (\%) and (b) corresponding forgetting measures on the S-CIFAR-10 dataset with different buffer sizes.
    Our method consistently improves accuracy and forgetting measures.
    }
    \label{fig:f_last_forgetting}
\end{figure}

We examine the proposed method in terms of the last accuracy and forgetting measurement. We apply the proposed method to two different baseline approaches.
We compute average forgetting at $T$th task as $\frac{1}{T-1}\sum_{t=1}^{T-1}\max_{j \in \{t,\cdots,T-1\}} a_{j}^{t} - a_{T}^{t}$ where $a_{j}^{t}$ means $t$th task accuracy at $j$ step.
Notably, higher forgetting means severe catastrophic forgetting of previous tasks.
\cref{fig:f_last_forgetting} shows that the view-batch model improves accuracy while addressing the forgetting problem simultaneously.

\section{Conclusion}
We propose the view-batch model, which is generally applicable to various continual learning scenarios.
Inspired by Ebbinghaus's forgetting curve theory, the proposed view-batch model optimizes the recall interval between retraining samples, improving neural networks' long-term memory.
The proposed method consists of two main components: replay and self-supervised learning.
The replay optimizes the recall interval for guaranteeing sufficient memory forgetting by grouping multiple views of a sample.
Also, the self-supervised learning ensures extensive learning for the multiple views using the one-to-many divergence loss.
Experimental results show that sufficient forgetting preserves long-term memory and ultimately improves continual learning accuracy.
We hope the proposed approach to continual learning can boost future research.
\\ \\
\noindent \textbf{Acknowledgment.}
This paper was supported in part by the ETRI Grant funded by Korean Government (Fundamental Technology Research for Human-Centric Autonomous Intelligent Systems) under Grant 24ZB1200, Artificial Intelligence Convergence Innovation Human Resources Development (IITP-2025-RS-2023-00255968), Artificial Intelligence Innovation Hub (RS-2021-II212068), and the NRF Grant (RS-2024-00356486).

{
    \small
    \bibliographystyle{ieeenat_fullname}
    \bibliography{main}
}

\newpage

\appendix

\section{Hyper-parameter Configuration}

\subsection{Baseline Method}
We provide hyper-parameters for each dataset used in this paper.
We describe our hyper-parameter settings for both from-scratch training and fine-tuning methods.
We train ResNet-18 from scratch while employing a fine-tuning method for ViT.
We use ViT-B/16, which is pre-trained on the ImageNet-21K dataset.
~\cref{tab:hyperparam_standard,tab:hyperparam_pretrain} show the key hyper-parameters used in our experiments.
We follow other specific hyper-parameters by baseline methods' for a fair comparison with them.
To provide more reliable experimental results, we report the mean and variance of three experimental results, each with different random seeds.

\begin{table}[!ht]
    \centering
    \footnotesize
    \caption{
    \textbf{Hyper-parameters for ResNet-18 backbone.}
    We show the key hyper-parameters used in this paper.
    }
    \renewcommand{\arraystretch}{1.1} 
    \setlength{\tabcolsep}{3pt}
    \label{tab:hyperparam_standard}
    \begin{tabular}{lcccc}
        \toprule
        Hyper-parameter & CIFAR-10 & CIFAR-100 & TinyImageNet & DomainNet \\ 
        \midrule
        Epochs & 50 & 170 & 100 & 50 \\
        Batch size & 32 & 128 & 32 & 128 \\ 
        Optimizer & SGD & SGD & SGD & SGD \\ 
        Learning rate & 0.03 - 0. 1& 0.03 - 0.1 & 0.03 - 0.1 & 0.03 - 0.1 \\ 
        LR scheduler & None & Multi-step & None & None \\ 
        Weight decay & None & None & None & None \\
        \midrule
        Network & \multicolumn{4}{c}{ResNet-18} \\
        \bottomrule
    \end{tabular}
\end{table}

\begin{table}[!ht]
    \centering
    \footnotesize
    \caption{
    \textbf{Hyper-parameters for ViT-B/16 backbone.}
    We describe the hyper-parameter settings for pre-trained networks.
    }
    \label{tab:hyperparam_pretrain}
    \renewcommand{\arraystretch}{1.1} 
    \setlength{\tabcolsep}{3pt}

    \begin{tabular}{lccc}
        \toprule
        Hyper-parameter & CIFAR-100 & ImageNet-R & DomainNet \\
        \midrule
        Epochs & 20 & 50 & 20 \\
        Batch size & 128 & 128 & 128 \\ 
        Optimizer & SGD & SGD & SGD \\ 
        Learning rate & 0.01 & 0.01 & 0.01 \\ 
        Milestones & 18 & 40 & 15 \\ 
        LR decay & 0.1 & 0.1 & 0.1 \\ 
        Weight decay & 5e-4 & 5e-4 & 5e-4 \\
        \midrule
        Network & \multicolumn{3}{c}{ViT-B/16} \\
        \bottomrule
    \end{tabular}
\end{table}

\subsection{View-Batch Model}

To ensure simplicity, we do not modify the baseline methods' hyper-parameters when applying our method to them.
Therefore, our method does not require hyper-parameter changes from baseline methods and does not need extra training or inference costs compared to baseline methods.
For augmentation type, we employ the widely-used auto-augmentation~\citep{cubuk2018autoaugment} method.
We do not change augmentation types for different datasets or methods for strict comparison.
This handcraft augmentation search will improve the network's performance, but we decided to stick to the same augmentation method to validate the proposed method's effect only without interference with augmentation.

\begin{table}
    \centering
    \scriptsize
    \renewcommand{\arraystretch}{0.95} 
    \setlength{\tabcolsep}{2pt}
    \begin{tabular}{lccccc}
        \toprule
        
        Method & Latency & RAM\textsubscript{CPU} & RAM\textsubscript{GPU}: forward & RAM\textsubscript{GPU}: backward \\ 
        \midrule
        
        Baseline & 29.5ms & 1.431GB & 0.210GB & 0.235GB \\ 
        +View-Batch-replay & 27.5ms & 1.422GB & 0.210GB & 0.235GB \\ 
        +View-Batch-SSL & 28.3ms & 1.425GB & 0.210GB & 0.235GB \\ 
        \bottomrule
    \end{tabular}
    \caption{\textbf{Experimental results on computing cost.} We use iCaRL as the baseline method on the S-CIFAR-10 dataset.}
    \label{tab:computing_cost}
\end{table}

\begin{table}
    \centering
    \footnotesize
    \renewcommand{\arraystretch}{0.95} 
    \setlength{\tabcolsep}{6pt}

     \begin{tabular}{lcrrrrr}
        \toprule

        \textbf{Method} & 
        \textbf{Task} &
        \specialcell{\textbf{RI=1}} & 
        \specialcell{\textbf{RI=2}} & 
        \specialcell{\textbf{RI=3}} &
        \specialcell{\textbf{RI=4}} &
        \specialcell{\textbf{RI=5}}
        \\
        \midrule
        
        \multirow{3}{*}{iCaRL} & CIL & 64.11 & 65.85 & 68.39 & 69.73 & 67.67 \\
        & TIL & 90.20 & 90.94 & 92.96 & 92.76 & 92.47 \\
        & Avg & 77.16 & 78.39 & 80.68 & 81.25 & 80.07 \\
        \midrule
        \multirow{3}{*}{DER++} & CIL & 61.67 & 65.79 & 66.99 & 61.68 & 63.44 \\
        & TIL & 90.61 & 93.57 & 94.30 & 93.82 & 94.42\\
        & Avg & 76.14 & 79.68 & 80.65 & 77.75& 78.93 \\

        \bottomrule
    \end{tabular}
    \caption{
    \textbf{Experimental results on recall interval.}
    We show the last top-1 accuracy varying the recall intervals (denoted as RI) on S-CIFAR-10.
    We use the ResNet-18 backbone in this experiment.
    }
    \label{tab:extended_recall_interval_analysis}
\end{table}

\begin{table*}[t]
    \centering
    \footnotesize

    \renewcommand{\arraystretch}{0.95} 
    \setlength{\tabcolsep}{6pt}
    
    \begin{tabular}{lcccccccccccc}
        \toprule
        \multirow{2}{*}{\textbf{Method}} & 
        \multicolumn{4}{c}{\textbf{5 Step}} &
        \multicolumn{4}{c}{\textbf{10 Step}} & 
        \multicolumn{4}{c}{\textbf{20 Step}} \\
        \cmidrule(lr){2-5}
        \cmidrule(lr){6-9}
        \cmidrule(lr){10-13}

        &
        \multicolumn{1}{c}{\textit{Avg}} &
        \multicolumn{1}{c}{\textit{$\Delta$}} &
        \multicolumn{1}{c}{\textit{Last}} & 
        \multicolumn{1}{c}{\textit{$\Delta$}} &
        \multicolumn{1}{c}{\textit{Avg}} &
        \multicolumn{1}{c}{\textit{$\Delta$}} &
        \multicolumn{1}{c}{\textit{Last}} & 
        \multicolumn{1}{c}{\textit{$\Delta$}} &
        \multicolumn{1}{c}{\textit{Avg}} &
        \multicolumn{1}{c}{\textit{$\Delta$}} &
        \multicolumn{1}{c}{\textit{Last}} &
        \multicolumn{1}{c}{\textit{$\Delta$}} \\
        \midrule

        DER  & 76.77 & - & 68.06 & - & 75.72 & - & 64.32 & - & 74.96 & - & 61.80 & - \\ 
        \rowcolor[gray]{.9}
        {\scriptsize \bfseries +replay} & 77.63 & {\scriptsize \bfseries \textcolor{OliveGreen}{+0.86}} & 69.21 & {\scriptsize \bfseries \textcolor{OliveGreen}{+1.25}} & 76.53 & {\scriptsize \bfseries \textcolor{OliveGreen}{+0.81}} & 65.49 & {\scriptsize \bfseries \textcolor{OliveGreen}{+1.17}} & 75.55 & {\scriptsize \bfseries \textcolor{OliveGreen}{+0.59}} & 62.65 & {\scriptsize \bfseries \textcolor{OliveGreen}{+0.85}} \\ 
        \rowcolor[gray]{.9}
        {\scriptsize \bfseries +self-supervised} & 78.60 & {\scriptsize \bfseries \textcolor{OliveGreen}{+1.83}} & 70.60 & {\scriptsize \bfseries \textcolor{OliveGreen}{+2.54}} & 78.12 & {\scriptsize \bfseries \textcolor{OliveGreen}{+2.40}} & 67.04 & {\scriptsize \bfseries \textcolor{OliveGreen}{+2.72}} & 76.95 & {\scriptsize \bfseries \textcolor{OliveGreen}{+1.99}} & 64.29 & {\scriptsize \bfseries \textcolor{OliveGreen}{+2.49}} \\ 
        \midrule
        
        TCIL & 77.33 & - & 69.48 & - & 76.33 & - & 65.66 & - & 74.32 & - & 62.54 & - \\ 
        \rowcolor[gray]{.9}
        {\scriptsize \bfseries +replay} & 78.42 & {\scriptsize \bfseries \textcolor{OliveGreen}{+1.09}} & 70.35 & {\scriptsize \bfseries \textcolor{OliveGreen}{+0.87}} & 77.02 &  {\scriptsize \bfseries \textcolor{OliveGreen}{+0.69}} & 67.71 & {\scriptsize \bfseries \textcolor{OliveGreen}{+2.05}} & 75.07 & {\scriptsize \bfseries \textcolor{OliveGreen}{+0.75}} & 63.93 & {\scriptsize \bfseries \textcolor{OliveGreen}{+1.39}} \\
        \rowcolor[gray]{.9}
        {\scriptsize \bfseries +self-supervised} & 79.23 &  {\scriptsize \bfseries \textcolor{OliveGreen}{+1.90}} & 71.23 & {\scriptsize \bfseries \textcolor{OliveGreen}{+1.75}} & 78.02 & {\scriptsize \bfseries \textcolor{OliveGreen}{+1.69}} & 68.14 &  {\scriptsize \bfseries \textcolor{OliveGreen}{+2.48}} & 76.83 & {\scriptsize \bfseries \textcolor{OliveGreen}{+2.51}} & 67.16 & {\scriptsize \bfseries \textcolor{OliveGreen}{+4.62}} \\   
        \bottomrule
    \end{tabular}

    \caption{\textbf{Experimental results on factor analysis.} 
    We showcase \textit{Avg} and \textit{Last} top-1 accuracy (\%) on the S-CIFAR-100 benchmark with three different class incremental steps. 
    ResNet-18 backbone is adopted for all networks.
    We follow the official implementation to reproduce the results of DER and TCIL.
    We demonstrate that our two main components significantly improve performance. 
    }
    \label{tab:factor_analysis}
\end{table*}

\section{Forgetting Curve Analysis Backgrounds}

In Section 1 of the manuscript, we illustrate the forgetting curve with different recall intervals.
We draw these forgetting curves based on spacing effect theory~\citep{ebbinghaus2013memory,cepeda2008spacing}.
In spacing effect theory, we estimate memory retention according to elapsed time and recall interval.
Specifically, in the forgetting curve theory~\citep{ebbinghaus2013memory}, human's memory retention $R$ could be defined as the function of the elapsed time $t$ from the initial learning experience as:
\begin{equation}
    R(t) = A(bt+1)^{-S},
\end{equation}
where $A$ is the first memory retention, $b$ denotes time scaling parameters, and $S$ represents the memory retention decay rate.
Obviously, we assume a higher decay rate of memory retention indicates faster forgetting.
Moreover, \citet{cepeda2008spacing} empirically proves that the decay rate depends on recall interval $I$.
Borrowing this empirical finding, the decay rate is defined as
\begin{equation}
    S=1+c(ln(I+1)-d)^2,
    \label{eq:decay_rate}
\end{equation}
where $d$ and $c$ are empirically determined parameters.
From \cref{eq:decay_rate}, we learn that the increasing recall interval improves the decay rate until some point $d$, then deteriorates it again.
\citet{wahlheim2014role} interpret this phenomenon as optimal recall interval mitigates a high decay rate due to adequate learning difficulty.
Finally, based on the given formula and theory, we illustrate different forgetting curves with varying recall intervals.

The degree of forgetting estimates the amount of memory neural networks forgets during recall intervals.
Namely, if the degree of forgetting is high, the neural networks significantly lose their memory before they relearn the same samples.
Not surprisingly, since excessive memory forgetting yields poor long-term memory retention in human learners~\citep{melton1970situation}, we employ our degree of forgetting as an empirical reason for the downward accuracy phenomenon in long-term recall interval in Figure 4 of the manuscript.

Specifically, we define the sequence of memory retention as $r_{0}, r_{1}, ..., r_{E-1}$.
Here, we denote $E$ for the number of total learning epochs and measure $r_i$ by evaluating neural networks on the current task at the end of each epoch and adopting their top-1 accuracy (\%) as a memory retention value.
Since we aim to quantify the degree of forgetting during recall interval, the variance of the memory retention values is used as our metric, calculating the averaged memory retention differences between the mean and individual retention values.
Leveraging the average memory retention difference, we define our degree of forgetting $\Delta_{r}$.
\begin{equation}
   \Delta_{r} = 
   \frac{1}{E-\overline{E}} 
   \sum_{i=\overline{E}}^{E} 
   ((\frac{1}{E-\overline{E}}\sum_{j=\overline{E}}^{E} r_j) - r_i)^2,
   \label{eq:memory_retention_difference}
\end{equation}
where we include memory retention values from the saturated epochs $\overline{E}$ due to high randomness in the early learning phase.
Consequently, we could evaluate the degree of forgetting of neural networks based on~\cref{eq:memory_retention_difference} in various CL scenarios.

\section{Additional Experimental Results}

\begin{table}
    \centering
    \footnotesize
    \renewcommand{\arraystretch}{0.95} 
    \setlength{\tabcolsep}{6pt}
    
    \begin{tabular}{lcc}
        \toprule
        
        \textbf{Method} & 
        \specialcell{\textbf{Degree of}\\\textbf{forgetting}} & 
        \specialcell{\textbf{Avg top-1}\\\textbf{accuracy (\%)}} \\
        \midrule
        
        Baseline & \textcolor{BrickRed}{\textbf{1.73}} & 76.33 \\ 
        VBM-C & \textcolor{BrickRed}{\textbf{6.20}} & 74.68 \\ 
        \rowcolor[gray]{0.9}
        VBM-S & \textcolor{OliveGreen}{\textbf{2.73}} & 78.11 \\ 
        \bottomrule
    \end{tabular}

    \caption{
    \textbf{Experimental results on View-Batch target.}
    We show the degree of forgetting and its average top-1 accuracy (\%) on the S-CIFAR-100 dataset varying the targets of view-batch model. 
    For the degree of forgetting, we represent prohibitive values with \textcolor{BrickRed}{\textbf{red}} and mark optimal values with \textcolor{OliveGreen}{\textbf{green}} color.
    }
    \label{tab:respacing_target_analysis}
\end{table}

\subsection{Results on Training Cost}
\cref{tab:computing_cost} analyzes the resources overhead in terms of latency (ms) and CPU and GPU RAM usage, which is measured by averaging three runs. The proposed method increases the minimal latency by 3\% to compute the KL divergence loss. Further, there is no additional resource usage for both CPU and GPU RAM when using our method. Therefore, we demonstrate that the proposed method can be utilized as the drop-in replacement approach.

\subsection{Results on  Recall Interval}

\cref{tab:extended_recall_interval_analysis} validates the various recall intervals on the S-CIFAR-10 datasets using two different baseline methods. We show that the optimal recall interval (\textit{i.e.}, \textbf{x3} or \textbf{x4}), which is found in Section \textcolor{BrickRed}{6} of the manuscript, generally works well in different baseline methods and different task types. This analysis demonstrates the proposed method's applicability in various continual learning scenarios.

\subsection{Results on Factor Analysis}

We extensively perform factor analysis in different CL scenarios.
\cref{tab:factor_analysis} demonstrate that the main components significantly improve respective baseline methods.
These experimental results reveal that the proposed components work consistently well across diverse CL scenarios.%

\subsection{Results on two types of View-Batch Model}

\begin{figure}[t]
    \centering

    \hfill
    \hfill
    \includegraphics[width=0.7\linewidth]{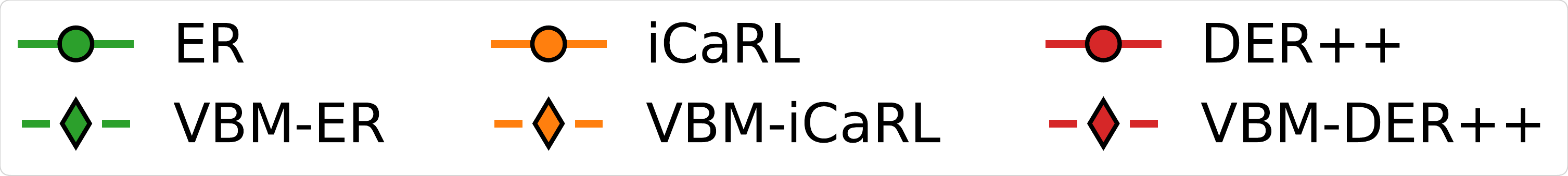}
    \hfill
    \hfill
    \hfill

    \hfill
    \hfill
    \begin{subfigure}[t]{0.45\linewidth}
        \centering        
        \includegraphics[width=\linewidth]{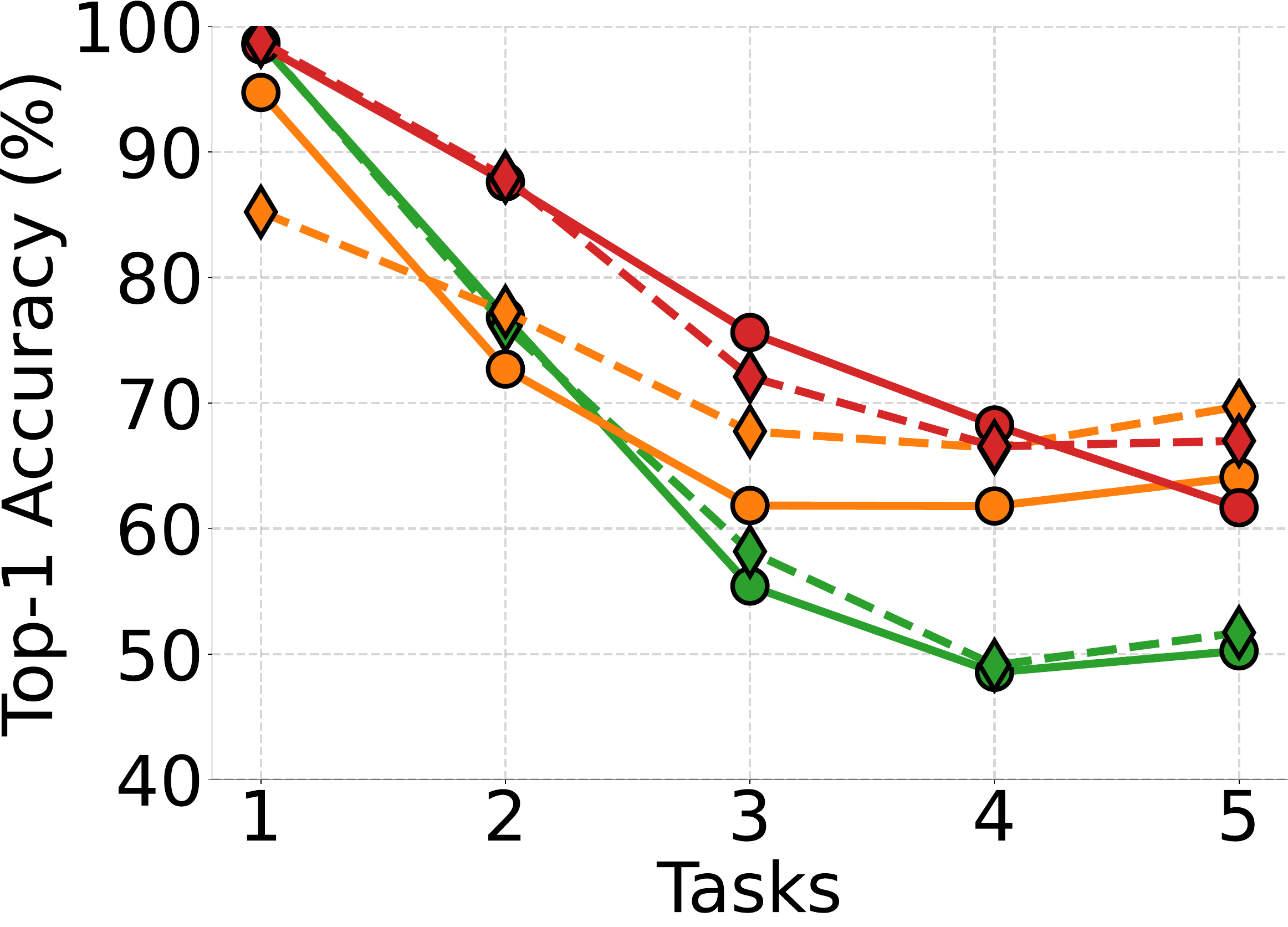}
        \caption{S-CIFAR-10}
        \label{fig:f_task_evolution_c10}
    \end{subfigure}
    \hfill
    \begin{subfigure}[t]{0.45\linewidth}
        \centering        
        \includegraphics[width=\linewidth]{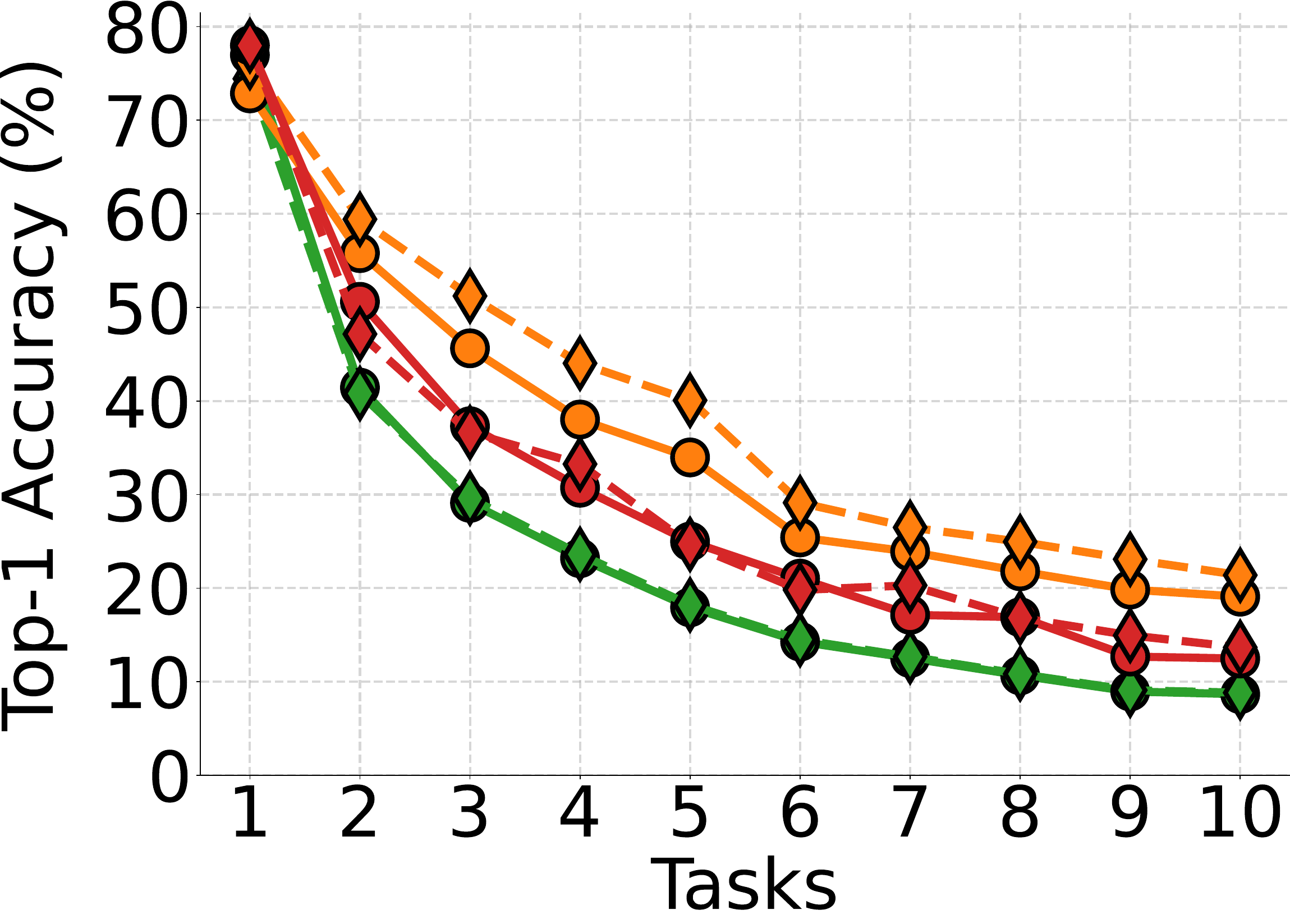}
        \caption{S-TinyImageNet}
        \label{fig:f_task_evolution_tinyimagenet}
    \end{subfigure}
    \hfill
    \hfill
    \hfill
    
    \caption{
    \textbf{Experimental results on evolving task.}
    We report class incremental accuracy at the end of each task, comparing the view-batch model to baseline methods. We provide details of the task evolution results in our repository.
    }
    \label{fig:f_task_evolution}
\end{figure}

We compare two types of the view-batch model, such as class- and sample-based approaches. 
While we augment a single sample to multiple views for constructing a {\ourssl}, one can also do this {\ourssl} construction in a class-based manner.
For the class-based view-batch (VBM-C) method, we constrain a single epoch to learn only specific class samples. If there are $C$ classes, we train only $C/N$ class samples N repeated times per epoch. This allows networks to learn the features of a specific class extensively in a single epoch.
However, \cref{tab:respacing_target_analysis} shows that the VBM-C over-escalates the degree of forgetting, leading to lower performance of continual learning.
On the other hand, our sample-based approach (VBM-S) increases memory retention fairly and achieves favorable performance compared to the class-based one.
We assume that our sample-based approach balances the recall interval and extensive learning.
Moreover, it indicates that it is more advantageous for self-supervised learning to learn all class samples in a single epoch.

\subsection{Results on Task Evolution}

We evaluate task evolution performances in the S-CIFAR-100 dataset as shown in~\cref{fig:f_task_evolution}.
In task evolution, we measure average class incremental accuracy at the end of each task and report all of them to compare our view-batch model and its respective baseline.
As a result, our view-batch model shows consistent performance improvements in all steps of various evaluation scenarios.

\end{document}